\documentclass[3p,times]{elsarticle}

\usepackage{tabularx}
\usepackage[ruled]{algorithm2e}
\usepackage{pdflscape}
\usepackage{subfig}
\usepackage{multirow}
\usepackage{footnote}
\usepackage{amssymb}
\usepackage[figuresright]{rotating}
\usepackage{color}
\usepackage{enumitem}

\newcommand{\etal}{et al. }
\newcommand\Tstrut{\rule{0pt}{3ex}}         
\newcommand\Bstrut{\rule[-1.5ex]{0pt}{0pt}}   

\begin{document}

\begin{frontmatter}




\title{Crowd Behavior Analysis: A Review where Physics meets Biology}


\author{Ven Jyn Kok}
\ead{venjyn.kok@siswa.um.edu.my}
\author{Mei Kuan Lim}
\ead{imeikuan@siswa.um.edu.my}
\author{Chee Seng Chan\corref{cor1}}
\ead{cs.chan@um.edu.my}

\address{Center of Image and Signal Processing,\\Faculty of Computer Science \& Information Technology,\\University of Malaya, 50603 Kuala Lumpur, Malaysia}

\cortext[cor1]{Corresponding author: Tel: +603-7967-6433}

\begin{abstract}
Although the traits emerged in a mass gathering are often non-deliberative, the act of mass impulse may lead to irrevocable crowd disasters. The two-fold increase of carnage in crowd since the past two decades has spurred significant advances in the field of computer vision, towards effective and proactive crowd surveillance. Computer vision studies related to crowd are observed to resonate with the understanding of the emergent behavior in physics (complex systems) and biology (animal swarm).  These studies, which are inspired by biology and physics, share surprisingly common insights, and interesting contradictions. However, this aspect of discussion has not been fully explored. Therefore, this survey  provides the readers with a review of the state-of-the-art methods in crowd behavior analysis from the physics and biologically inspired perspectives. We provide insights and comprehensive discussions for a broader understanding of the underlying prospect of blending physics and biology studies in computer vision.
\end{abstract}

\begin{keyword}
crowd behavior analysis, biologically-inspired, physics-inspired, computer vision, survey
\end{keyword}

\end{frontmatter}


\section{Introduction}
\label{Introduction}

\begin{quote}
\emph{``The one who follows the crowd will usually go no further than the crowd; the one who walks alone is likely to find herself in places no one has ever been before", Albert Einstein}. 
\end{quote}

While this quote is lived by many, this paper is motivated by the contrary. Our work is based on the notion that literally, \emph{one who follows the crowd will surpass solitary individual, and together with the crowd, `venture beyond places' where no lone individual is capable of venturing to}; a phenomenon known as the emergent behavior. Emergent behavior arises in a swarm or crowd with certain class of entities (e.g. insects, human, animals, etc.); whereby, each entity is self-organized and together they portray a complex and coordinated collective behavior. The essence of the emergent behavior is based on a simple rule of thumb, where entities engage with one another using basic interactions. This in turn heightens ones' sense of responsiveness to the surrounding, and instantaneously brings them closer to their goal. What makes it interesting is that, this resultant phenomenon is not possible to be achieved by solo individuals. 

Over the past years, the biologists have observed the emergent of collective behaviors in organism, insects and animals and were constantly investigating the underlying mechanism that allows unity in a swarm \cite{parrish1999complexity,Camazine01,Couzin03,Krause02}. For example, a school of fish that swims together and yet not colliding with each other, or a flock of starlings steering in the air with the uncanny synchronization. The slime mold that exist as a single-cell organism, congregate to form multicellular when food supplies is scarce, working in tandem to search for the shortest path to food source. Another well-known example is the foraging activity of a colony of ants. Although each ant follows a set of simple rules, the colony as a whole, acts in a sophisticated way that increases its foraging efficiency \cite{Dorigo08}. Fascinatingly, this similar behavior has been observed in human crowds as well. Amongst the early works that were motivated by the emergent behavior in human crowds was the concept of the `mind' by Le Bon in~\cite{Lebon1986} which stated that, when individuals in a crowd gather and coalesce, a new distillation of traits emerged. He referred to the emergent behavior as collective `unconsciousness' that robs every individual member of their opinions, values and beliefs. He put forward that the emergent behavior is very subtle and ignorant to each individual, yet, is capable of forming intriguing collective `group mind' that works wonders. This phenomenon can be seen commonly in a crowded scene. For instance, when two flows of people moving in the reverse directions, a uniform walking lanes for each direction would be formed spontaneously although there is no communication amongst the individuals in the crowd.

In existing literature, the dynamics of human crowd are often studied through analogies with theories in physics and biology. The idea of relating the motion of crowd with fluid, liquid or electrons in aerodynamics, hydrodynamics or continuum mechanics respectively, has generated many research in crowd analysis since the past years \cite{Henderson74,Helbing07}. Accordingly, physics-inspired studies assume that the individual in a crowd tends to follow the dominant flow of the crowd and thus, the motion of highly dense crowd resembles fluid. Hence, theories and methods in fluid mechanics are adopted to comprehend the flow of human crowd. In another physics-inspired example, the kinetic theory of gases is applied to model the sparse and random interaction forces amongst individuals in a crowd.  On the contrary, from the biology point of view, individuals in a crowd resemble the entities in a swarm. Each individual in the swarm exhibits diverse interaction forces towards achieving the final goal, which is apparently common amongst members in the swarm \cite{Helbing95,Moussaid11}. For example, the motion of individuals in a train station, where everyone is moving with different pace towards the common exit region, or the diverse motion of individuals finding their ways to the boarding area. 

Nevertheless, there is no clear distinction between the approaches inspired by the two sciences; physics and biology. Instead, we observe that some terminologies or notions from both approaches share interestingly similar understanding and perspective, while holding on to some minor differences. The studies of the human crowd behavior from the perspectives of the two sciences drawn into the field of computer vision is a new and rapidly developing study \cite{Spears12}. It is predominantly deemed as a notion for crowd behavior analysis to enhance and assist the analysis of visual crowd surveillance, which aims to imitate the human visual perception. The capability to emulate human visual perception allows the development of practical systems that provide meaningful and concise description of crowd behavior, to better assist human in crowd surveillance, which is the focal interest of this study.

\subsection{Comparisons with Previous Reviews}
\label{RelatedWork}

Although there have been great interest and a large number of methods have been developed for crowd analysis in general, there are limited comprehensive reviews which focused on crowd behavior understanding~\cite{Thida13}. Most existing survey papers~\cite{Zhan08,Junior10,sjarif2011detection,loy2013crowd,Thida13,li14} focuses on the computer vision techniques and review the essential features required for application specific crowd analysis. To the best of our knowledge, none of the aforementioned reviews provide in-depth discussion from the perspectives of physics or biologically-inspired approaches in the context of crowd behavior analysis.

The closest attempt to bridge the studies between physics and biology in the context of crowd behavior understanding was by Hughes \cite{Hughes03}. His work emphasizes on the key distinctions between physics and the actual crowd. Although the discussion was focused only on crowd modeling from the physics perspective, the concept that described aptly the `thinking' component of fluids spurred thought that the interactions between individuals in a crowd is far more complex than particles in fluid. This coincides with the understanding of crowd motion in biology. Another work in \cite{Leggett04} categorized the state-of-the-art methods in crowd simulation into three broad approaches which include i) fluids, ii) cellular automata and iii) particles. He suggested the classification of existing work without discussing much on the underlying motives and attributes between these categories. In addition to the 3 broad categories proposed by Leggett in \cite{Leggett04}, Zhan \etal\cite{Zhan08} reviewed approaches to infer crowd events by further dividing the `particles' category into agent and nature-based models; leading to 4 categories of crowd models from the non-vision approaches. This includes i) physics-inspired, ii) agent-based, iii) cellular automation and iv) nature-based. While their work acknowledged the advantages of integrating the non-vision models with computer vision methods for crowd analysis, the in-depth discussion on the different non-vision models from the physics and biology perspectives is lacking. Thida \etal in \cite{Thida13} presented a review with systematic comparisons of the state-of-the-art methods in crowd analysis, where the merits and weaknesses of various approaches were discussed comprehensively. Their work is based on the three distinct philosophies for modeling a crowd by Alexiadis \etal in \cite{FHWA04}, where crowd models are categorized as microscopic, mesoscopic and macroscopic. The microscopic model deals with the crowd as discrete individuals while the macroscopic model treats the crowd as a unit. The mesoscopic model combines the properties of the former two models, that is, the microscopic state of pedestrians are maintained with an addition of the general view of crowd. Yet, the gap between the two approaches has not been discussed clearly.

Other papers are more specific towards understanding crowd behavior, disregarding the point of whether the different methods of analysis are inspired by the studies from physics or biology. Each of the works provides critical outlook of existing literature pertaining to the different aspects of crowd analysis and serves as a reference point to all computer vision practitioners in the domain. However, we observed that a great deal of them are focused on \textbf{physics-inspired} approaches. Helbing \etal in \cite{Helbing072} discussed their analysis on using density and pressure attributes to infer two new phenomena in crowd; the stop-and-go and turbulent flows. Their discussions are highly influenced by physics and provide readers with insights to where and when accidents tend to occur in crowded scenes, and on how the proper management of crowd can ensure prevention of crowd disasters. In another review that is based on the notion that individuals in crowds behave in ways like particles in the fluid is by Moore \etal\cite{Moore11}. Their work adopted the concept of scale in hydrodynamics (the study of liquid in motion) as opposed to the common adaptation of aerodynamics (the study of gaseous or air in motion). The main difference between the two is that in the former, the interaction forces between individuals in the crowd tend to dominate the motion of the individuals, while in the latter, the interactions between individuals are few and random motion is most likely to dominate the crowd behavior.  In a more recent review, Jo \etal\cite{jo2013review} briefly highlighted the difference between physics-based and physics-inspired methods. Accordingly, physics-based methods are rooted in fundamental physic ideas whereas the latter are inspired by the laws of physics. In \cite{Moussaid11,Moussaid112}, the limitations of existing physics-inspired models to describe pedestrian behaviors and crowd disasters are discussed comprehensively. This includes the difficulty to capture the complexity of crowd behaviors using a single model and the insufficiency of current models in understanding the interactions between individuals and their environment.  Thus, they introduced the integration of \textbf{cognitive science and physics} for a more holistic solution. Some examples of the heuristic rules which is derived from the natural cognitive of human include the assumptions that an individual tend to move towards a possible entry or exit, and that an individual is very likely to move its motion according to his or her gaze angle. Interestingly, the introduction of such simple rules adheres to the concept of emergent behavior, where the collective dynamics of a social system with many interacting individuals can be modeled through simple rules. A more comprehensive review of physics-inspired crowd models covering the 3 main aspects of crowd motion pattern segmentation, crowd behavior recognition and anomaly detection can be found in \cite{li14}. While this review provide broad discussion on existing models, algorithms and evaluation protocols of research in crowd, the outlook of computer vision approaches from the perspectives of physics and biology remains unstated. Other relevant researches include the study on crowd dynamics and how the different dynamics of crowd can lead to the various issues in crowd safety by Johansson \etal\cite{Johansson11}, the modeling of crowd dynamics from the viewpoint of \textbf{mathematics} \cite{Bellomo2012}, the analysis of human behaviors from the perspectives of \textbf{social signal processing} \cite{CristaniNeuro2013}, the study of crowd dynamics from the \textbf{psychology} perspective by Reicher in \cite{Reicher01}, the underlying rules that lead to collective behaviors for group intelligence problem-solving by Fisher \cite{Fisher09} and the comprehensive review on the basic laws of \textbf{physics and mathematics} that describe collective motion which leads to the emergent behavior in groups of animals or humans \cite{Vicsek12}. 

To the best of the authors' knowledge, there is no review on biologically-inspired algorithms for crowd analysis in computer vision. This is rather surprising, given the plethora of methods that apply biological concepts for crowd analysis today \cite{Musse97,HelbingHerding2000,Lin07,Krause10}.  A summary of the existing surveys are presented in Table \ref{TableSummary} and \ref{TableSummaryDiscipline}.


\begin{landscape}
\begin{table}[htbp]
\centering
\caption{Summarization of the review papers on crowd behavior analysis. }
\begin{center}
\label{TableSummary}
\resizebox{21.5cm}{!}{
    \begin{tabular}{c p{6cm} p{6.5cm} p{13.5cm}c}
    \hline
    Paper \Tstrut\Bstrut       & Author       															& Title                                				& Description                                                   & Year \\ \hline \hline \\
    \cite{Hughes03}            & R. L. Hughes 															& The flow of human crowds             				& Presented the notion that crowd is a continuum with the capability to think. The crowd motion can be represented as `thinking fluids'.  & 2003 \\ \\
    \cite{Leggett04}           & R. Leggett   															& Real-time crowd simulation: A review 				& Reviewed the research on real-time crowd simulation, focusing on the approaches that modeled pedestrian as fluid, cellular automata or particles. Also, describe the implementation of CrowdSim model to simulate crowd motion. & 2004 \\ \\
    \cite{Zhan08}              & B. Zhan, D. N. Monekosso, P. Remagnino, S. A. Velastin, \& L. Q. Xu 	& Crowd analysis: A survey 							& Surveys crowd analysis methods in computer vision, covering approaches of crowd density estimation, crowd tracking as well as pedestrian and crowd recognition. Discusses crowd model from the perspective of other research discipline and the potential to integrate with computer vision for crowd modeling and events inferencing.	& 2008 \\ \\
    \cite{Fisher09}			   & L. Fisher 																& The Perfect Swarm: The Science of Complexity in Everyday Life & Explores the collective behavior emerged from a set of very simple rules of interaction between neighbouring entities (specifically insects such as locusts, bees and ants). Focus on the development of group intelligence in human crowd to solve complex problems. & 2009 \\ \\
    \cite{Junior10}            & J. C. S. Jacques Junior, S. Raupp Musse, \& C. R. Jung 															& Crowd analysis using computer vision techniques 	& Overview of computer vision techniques for crowd analysis, specifically on people tracking, crowd density estimation, event detection, validation and simulation. Also, describe the correlation between crowd simulation and analysis to deal with challenges in crowd analysis.	& 2010 \\ \\
    \cite{Moore11}             & B. E. Moore, S. Ali, R. Mehran, \& M. Shah 							& Visual crowd surveillance through a hydrodynamics lens & Reviewed hydrodynamics-based techniques to model the interaction forces between individuals in crowd for crowd analysis in visual surveillance of high-density crowd.	& 2011 \\ \\
    \cite{sjarif2011detection} & N. N. A. Sjarif, S. M. Shamsuddin \& S. Z. Hashim						& Detection of abnormal behaviors in crowd scene: A review 	& Presented the advances in the studies of detecting abnormal behavior in crowded scenes from 2000 till 2010. & 2011 \\ \\
    \cite{Vicsek12}            & T. Vicsek \& A. Zafeiris  												& Collective motion       							& Reviewed the observation and basic laws of collective motion which is on the borderline of several scientific disciplines. & 2012 \\ \\
    \cite{Thida13}             & M. Thida, Y. L. Yong, P. Climent-Perez, H. L. Eng \& P. Remagnino 		& A literature review on video analytics of crowded scenes & Reviewed on the state-of-the-art approaches in automatic crowd video analysis, by emphasizing on the macroscopic modeling, microscopic modeling and crowd event detection. & 2013 \\ \\
    \cite{jo2013review}        & H. Jo, K. Chug \& R. J. Sethi 											& A review of physics-based methods for group and crowd analysis in computer vision & Presented a review of the physic-based approach for group and crowd analysis in computer vision. & 2013 \\ \\
    \cite{loy2013crowd}        & C. C. Loy, K. Chen, S. Gong \& T. Xiang 								& Crowd counting and profiling: Methodology and evaluation & Reviewed the state-of-the-art approach for video imagery based crowd counting with emphasis on the methodologies and systematic evaluation of different techniques. & 2013 \\ \\
    \cite{li14}                & T. Li, H. Chang, M. Wang, B. Ni, R. Hong \& S. Yan 					& Crowded scene analysis: A survey  & Focused on the techniques for crowded scene analysis from 2010 onward, covering the task of motion pattern recognition, crowd behavior recognition and anomaly detection. Outline the available datasets for performance evaluation.	& 2014 \\ \\ \hline
    \end{tabular}
    }
\end{center}
\end{table}
\end{landscape}

\begin{landscape}
\begin{table}[htp]
\caption{The disciplines and criteria emphasized in the review papers on crowd behavior analysis. Note that item without check-mark indicates that the topic is not discussed comprehensively in the respective review paper, but may have been mentioned intrinsically in the context.}
\begin{center}
\label{TableSummaryDiscipline}
\resizebox{21.5cm}{!}{
	\begin{tabular}{cl|cccc|cccccc}
	\multirow{2}{*}{Year} 		& \multirow{2}{*}{Paper}  	& \multicolumn{4}{c|}{Discipline of Discussion} 			   				& \multicolumn{6}{c}{Aspect of Discussion}                                       \\ \cline{3-12}
	       						&						 	& Physic \Tstrut\Bstrut    	& Biology     & Computer Vision  & Others 		& Crowd Theory  & Feature     & Model    		& Comparative Comparison & Application & Dataset \\\hline\hline
	       2003 \Tstrut\Bstrut  & \cite{Hughes03}         	& \checkmark   				& -           & -                & -			& \checkmark	& -           & \checkmark  	& -            			& \checkmark  & -			\\ 
	       2004 \Tstrut\Bstrut  & \cite{Leggett04}        	& \checkmark 				& \checkmark  & -                & \checkmark	& -				& -           & \checkmark		& -            			& \checkmark  &	-			\\ 
	       2008 \Tstrut\Bstrut  & \cite{Zhan08}           	& \checkmark				& \checkmark  & \checkmark       & \checkmark	& -      		& \checkmark  & \checkmark		& -            			& \checkmark  &	-			\\ 
	       2009 \Tstrut\Bstrut  & \cite{Fisher09}         	& \checkmark 				& \checkmark  & -                & \checkmark	& \checkmark	& -           & -        		& -            			& -      	  & -			\\ 
	       2010 \Tstrut\Bstrut  & \cite{Junior10}         	& -          				& -           & \checkmark       & -			& -       		& \checkmark  & \checkmark		& -            			& \checkmark  & -			\\ 
	       2011 \Tstrut\Bstrut  & \cite{Moore11}          	& \checkmark				& -           & \checkmark       & -			& -       		& -           & \checkmark		& -           			& \checkmark  & -			\\ 
	       2011 \Tstrut\Bstrut  & \cite{sjarif2011detection}& -          				& -           & \checkmark       & -			& -       		& \checkmark  & \checkmark		& -            			& \checkmark  & \checkmark	\\ 
	       2012 \Tstrut\Bstrut  & \cite{Vicsek12}         	& \checkmark				& \checkmark  & -                & \checkmark	& \checkmark	& -           & \checkmark		& -            			& -      	  & -			\\ 
	       2013 \Tstrut\Bstrut  & \cite{Thida13}          	& -			  				& -           & \checkmark       & -			& -       		& \checkmark  & \checkmark		& -            			& \checkmark  & \checkmark	\\ 
	       2013 \Tstrut\Bstrut  & \cite{jo2013review}     	& \checkmark 				& -           & \checkmark       & -			& -       		& -       	  & \checkmark		& -            			& -      	  & -			\\ 
	       2013 \Tstrut\Bstrut  & \cite{loy2013crowd}     	& -          				& -           & \checkmark       & -			& -       		& \checkmark  & \checkmark      & \checkmark   			& \checkmark  & \checkmark 	\\ 
	       2014 \Tstrut\Bstrut  & \cite{li14}             	& \checkmark  				& -           & \checkmark    	 & - 			& -        		& \checkmark  & \checkmark      & \checkmark			& \checkmark  & \checkmark  \\  \hline	      	       	       
	\end{tabular}
	}
\end{center}
\end{table}
\end{landscape}

\subsection{Motivation and Contributions}

\begin{figure}[!h]
\centering
\includegraphics[height=0.39\linewidth, width=0.89\linewidth]{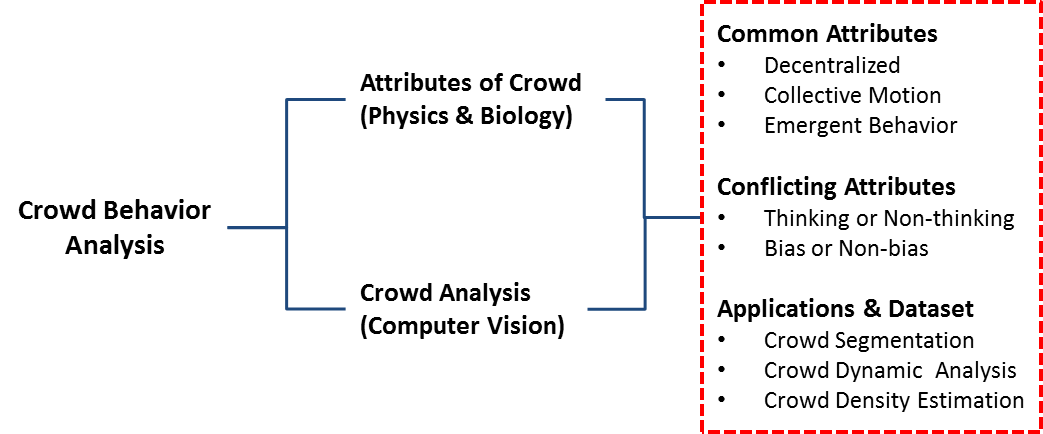}
\caption{The overall organization of this review.}
\label{fig:taxonomy}
\end{figure}

A thorough exploration on the current literature provides a broader outlook on the potentials of computer vision to cut across disciplines, especially in the two areas of sciences; physics and biology, to further enhance the efficiency of video surveillance of crowded scenes.  Therefore, in this paper, we focus primarily on the attributes of crowd behavior from the study of physics and biology in computer vision. We discuss existing computer vision solutions that integrate these attributes for crowd behavior analysis. Specifically, this paper provides comprehensive review of the state-of-the-art computer vision methods in crowd behavior analysis from the physics and biologically inspired perspectives. We put forward the underlying prospects of leveraging the studies from the two sciences and bringing together different disciplines in the hope of, and envisaging computer vision to the next level. As shown in Figure \ref{fig:taxonomy}, to ease understanding and improve readability, we begin with comprehensive introduction of the key attributes; common and conflicting attributes of crowd behavior from the perspectives of the two sciences. This is then followed by detailed discussion on the state-of-the-art crowd analysis applications and the commonly used benchmark dataset in each respective area. Here, we discuss the applications of computer vision in the perspective of their key attributes. The applications are divided into three common tasks in computer vision generally: i) crowd segmentation, ii) crowd dynamic analysis and iii) crowd density estimation. It is important to note that this study seeks to initiate the outlook of computer vision solutions from the viewpoint of physics and biology. This is as opposed to previous surveys, especially computer vision related reviews such as in~\cite{li14}, that focuses on computer vision techniques (e.g. feature representation and model learning) in crowd analysis.  Ideally, this study hopes to spark interest in integrating multiple disciplines for the advancement of crowd analysis in computer vision.

The remainder of this paper is organized as follows: Section \ref{IntroTerms} introduce the terms swarm, crowd and their relation to computer vision. In Section \ref{GeneralAttributes} and \ref{ContradictingAttributes}, we outline the various attributes of crowd and discuss their similarities and differences with modeled crowd behavior. Section \ref{Applications} presents a summary of the state-of-the-art computer vision applications, particularly in the branch of crowd behavior analysis. In addition, we discuss these applications with regards to the shared attributes between the two sciences to ease understanding of the concept of emergent behavior. Section \ref{Prospect} provides the opinions of the authors with regards to the forthcoming of a multidisciplinary crowd behavior analysis. Finally, in Section \ref{Conclusion}, we would conclude with our insights on the potential of spanning the distance between the physics and biologically inspired approaches in computer vision, for crowd behavior analysis and understanding in particular.

\section{Swarm, Crowd and Computer Vision - the Beginnings}
\label{IntroTerms}

\subsection{What is a swarm or crowd?}
\begin{figure}[!h]
\centering
\subfloat[A colony of ants displays collectively intelligent behavior when foraging for food.]{\includegraphics[height=0.2\linewidth, width=0.3\linewidth]{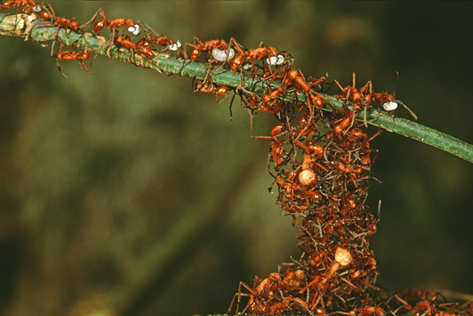}}
\qquad
\subfloat[The mesmerising behavior of large flock of starlings when they fly together.]{\includegraphics[height=0.2\linewidth, width=0.3\linewidth]{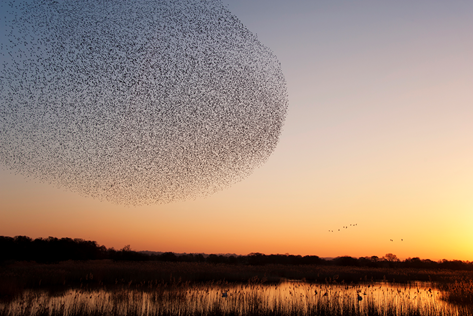}}
\qquad
\subfloat[The simulation of smoke using fluid dynamic models.]{\includegraphics[height=0.2\linewidth, width=0.28\linewidth]{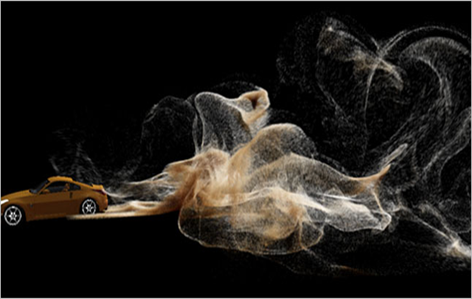}}
\caption{Examples of swarm in the two sciences; biology and physics.}
\label{fig:Swarm}
\end{figure}

\begin{figure}[!h]
\centering
\subfloat[Separation]{\includegraphics[height=0.22\linewidth, width=0.22\linewidth]{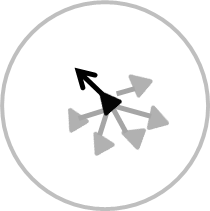}}
\qquad
\subfloat[Alignment]{\includegraphics[height=0.22\linewidth, width=0.22\linewidth]{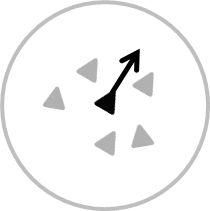}}
\qquad
\subfloat[Cohesion]{\includegraphics[height=0.22\linewidth, width=0.22\linewidth]{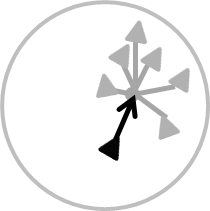}}
\caption{The (simple) interactions between agents in a swarm according to the Boids model gives rise to the (complex) emergent behavior.}
\label{fig:Boids}
\end{figure}

A swarm or crowd is generally referred to as a collection of spatially proximate class of entities. Most commonly, the entities are associated to insects such as bees and ants, thus elucidating the enormous research efforts in swarm intelligence that are motivated by biology. For example, the well-known Ant Colony Optimization algorithm that was inspired by the foraging behavior of a colony of ants and the Particle Swarm Optimization algorithm which simulates the synchronized movement of a flock of birds. Beekman \etal\cite{Beekman08} advocate that swarm intelligence is biology. On the other hand, the physicists have for a long time used the term swarm to refer to particles or electrons to describe transport equations and fluid flows \cite{Brinkman49,Dutton75,Dote80}. Examples of swarm are as shown in Figure \ref{fig:Swarm}. For decades, the biologists and physicists have been studying the behaviors of social animals to investigate the underlying mechanism that allows unity in a swarm. Only in the late-80s that computer scientists begin to discover the potentials of the swarm intelligence and proposed scientific insights of these algorithms into the different applications in the field of computer vision such as robotics \cite{Beni89,Muniganti10}, optimization \cite{Blum08,Krause13} object tracking \cite{Thida09,Lim13}. In fact, the great leap forward was made by Reynolds \cite{Reynolds1987}, where he created a computer model known as Boids, comprising a large group of virtual agents that mimicked the coordinated movement of a flock of birds. This simulation applied simple rules to control the steering behaviors of its agent: separation (keeping some distance from other agents), alignment (move at a velocity that matches with local flock mates) and cohesion (move towards the average position of local flock mates). The graphical representation of the rules applied by Boids is as shown in Figure \ref{fig:Boids}. Since then, great strides have been taken to investigate if the emergent of complex behaviors is indeed caused by simple rules and interactions amongst individuals in a large group, mainly through simulations \cite{Helbing95}. In the same way, the simulated model by Helbing and Molnar in \cite{Helbing95}, and the Boids has sparked a wide interest in the adaptations of the emergent behaviors to solve complex computer vision problems.  

\subsection{What is the role of computer vision in crowd analysis?}

\begin{figure}[!h]
\centering
\subfloat[Crowd in the pilgrimage or Hajj scene.]{\includegraphics[height=0.2\linewidth, width=0.3\linewidth]{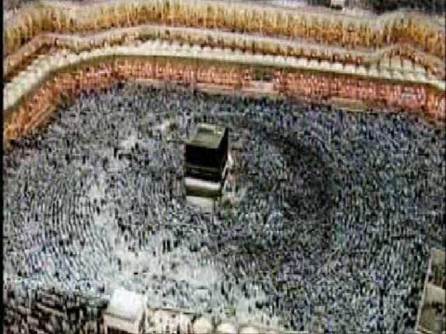}}
\qquad
\subfloat[Crowd in a train station.]{\includegraphics[height=0.2\linewidth, width=0.3\linewidth]{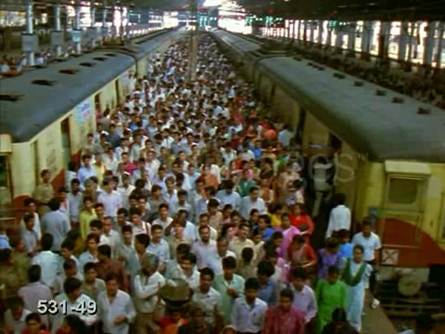}}
\qquad
\subfloat[Crowd of spectators in a stadium.]{\includegraphics[height=0.2\linewidth, width=0.3\linewidth]{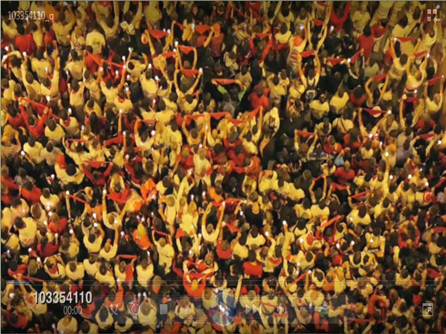}}
\caption{Example scenarios of highly dense crowd scenes which are taken from commonly used benchmark dataset in computer vision.}
\label{fig:CrowdA}
\end{figure}

\begin{figure}[!h]
\centering
\subfloat[Hillsborough disaster]{\includegraphics[height=0.2\linewidth, width=0.3\linewidth]{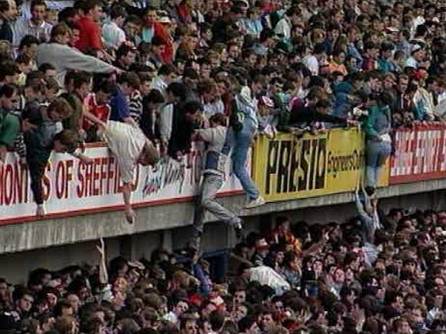}}
\qquad
\subfloat[Philsports stadium disaster]{\includegraphics[height=0.2\linewidth, width=0.3\linewidth]{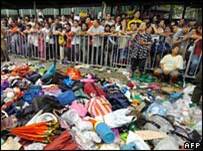}}
\qquad
\subfloat[Love Parade disaster]{\includegraphics[height=0.2\linewidth, width=0.3\linewidth]{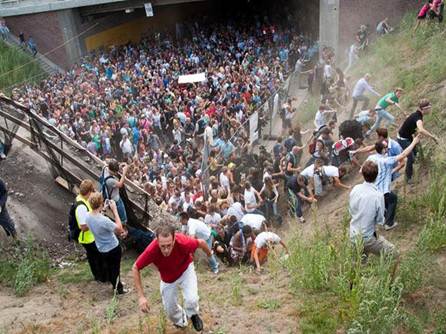}}
\caption{Some cases of crowd disasters at mass gathering events.}
\label{fig:CrowdB}
\end{figure}

One particular field of interest in public security involves visual surveillance of mass gathering of people, such as during public assemblies (e.g. music festivals, religious events) and demonstrations (e.g. strikes, protests) as illustrated in Figure~\ref{fig:CrowdA}. The security of mass crowd during public events has always been of high concern to relevant authorities due to the dynamics and degeneration risk. Any crowded environment has high tendency to plunge into a panic atmosphere given physical stress (i.e. overcrowding) or sudden  external pressure (i.e. shootings, fire), where the consequences are often devastating. Various examples from historical incidents have shown how things can easily get out of control when mass of people come together during big events. Some examples of disasters that have happened in the past are as shown in Figure~\ref{fig:CrowdB}. 

One must understand that in crowded scenes, where crowds of hundreds or even thousands gather, video monitoring is a daunting task. Often, incidents within a crowd went unnoticed due to the inherent limitations from depending solely on manual monitoring by CCTV operators. The limitations are commonly due to i) sheer number of screens to be monitored, ii) boredom and human fatigue, iii) distractions and interferences, and iv) the complexity and uncertainty of human behavior. In most scenarios, the consequences of not being alert of overcrowding and fail detection of suspicious activities may ultimately lead to unfavorable incidents which are irreversible and catastrophic. Table \ref{table:TableMassDisaster} lists some cases of crowd disasters at mass gathering events. Still in \cite{StillTableDisaster2014} and Soomaroo and Murray in \cite{LeeTableDisaster2012} provide comprehensive summary of crowd disaster.

\begin{figure}[!h]
\centering
\subfloat[Example of the flow direction for the pilgrimage sequence, where the red arrows denote motion to the right while blue arrows indicate motion towards the left of the image. The flow field which is estimated using optical flow is then refined using the Lagrangian Particle Dynamics approach for crowd flow segmentation \protect \cite{Ali2007}.]{\includegraphics[height=0.2\linewidth, width=0.3\linewidth]{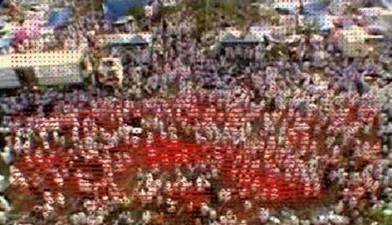}}
\qquad
\subfloat[Crowd density estimation using a combination of features such as head detection and texture elements, which is then fed into a Markov Random Field for consistent density estimation. In this image the ground truth number of individual is 1567 while the estimated number as reported in \protect \cite{Haroon2013} is 1590.]{\includegraphics[height=0.2\linewidth, width=0.3\linewidth]{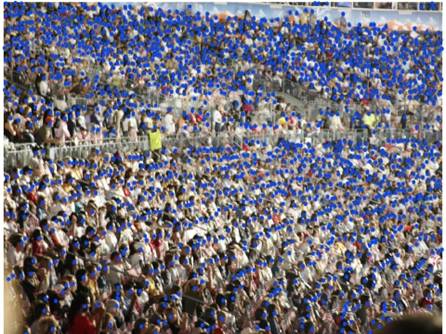}}
\qquad
\subfloat[The chaotic dynamics representation of trajectories are extracted and quantified in order to identify anomaly. In this scenario, the yellow bounding boxes indicate abnormal dancing behavior of individuals in the scene, where the learned normal activity is clapping behavior \protect \cite{WuChaotic2010}. ]{\includegraphics[height=0.2\linewidth, width=0.3\linewidth]{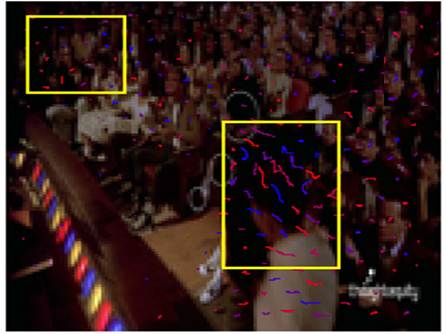}}
\caption{Example of computer vision applications in crowd behavior analysis.}
\label{fig:CrowdC}
\end{figure}

Carnage in crowd happens for a variety of reasons and have seen a two-fold increase in the past two decades \cite{JamesDisasterPub2010,NgaiIndia2013}. The aftermath investigations surrounding most of the crowd disasters conclude that there were missed opportunities to use technology for crowd behavior understanding to achieve proactive surveillance \cite{JoshuaJainCaseStudy2013}. Hence, the recent years have seen significant advances in using computers and technologies, specifically in the field of computer vision, to assist humans in the task of video monitoring for a more efficient and proactive crowd surveillance (as shown in Figure~\ref{fig:CrowdC}). This survey draws a bead on physics and biology inspired methods for crowd analysis, which is a challenging research topic in computer vision.

\begin{landscape}
\begin{table} 
\centering
\caption{Examples of crowd disasters at mass events. }
\label{table:TableMassDisaster}
\resizebox{21.5cm}{!}{
\begin{tabular}{|p{1.6cm}|p{5cm}|p{13cm}|p{4.6cm}|c|}
\hline 
\textbf{Date}\Tstrut\Bstrut & \textbf{Event - Place} & \textbf{Description} & \textbf{Casualties} & \textbf{Reference} \\
\hline
\hline
Jan 1971\Tstrut\Bstrut & Ibrox disaster (football match) - Glasgow, UK & Crush between fans entering and exiting.  & 
66 deaths, 140 injured & \cite{PopplewellReport1986} \\
\hline
Feb 1981\Tstrut\Bstrut & Nightclub fire - Ireland & Fire was started deliberately in the alcove.  & 
48 deaths, 128 injured & \cite{TribunalDublinFire1981} \\
\hline
Apr 1989\Tstrut\Bstrut & Hillsborough disaster (football match)- Sheffield, UK & Crush due to overcrowding surge against barrier.  & 
96 deaths, 766 injured & \cite{LordTaylorHillsborough1989} \\
\hline
Jul 1990\Tstrut\Bstrut & The Hajj disaster – Mecca, Saudi Arabia & Crush caused by lack of directional flow of pilgrims and crowd control in the tunnel.  & 
1426 deaths, no data is available for injured & \cite{AlamriHajj2014} \\
\hline
Jan 1991\Tstrut\Bstrut & Orkney stadium disaster - South Africa & Crush when fans panic and try to escape from brawls that break out in the grandstand.  & 
40 deaths, 50 injured & \cite{DarbySoccerDisaster2005} \\
\hline
Jan 1993\Tstrut\Bstrut & New year’s eve stampede – Lan Kwai Fong, Hong Kong & Slip and fall which leads to more and more people deprived of footing and fell; piling on top of another.  & 
21 deaths, no data is available for injured & \cite{WuHongKong2011} \\
\hline
May 1994\Tstrut\Bstrut &  The Hajj disaster – Mecca, Saudi Arabia & Progressive crowd collapse caused by the sheer number of pilgrimages.  & 
266 deaths, 98 injured & \cite{GadelHajj2008} \\
\hline
Jul 2001\Tstrut\Bstrut &  Akashi pedestrian bridge accident - Akashi Japan & Crush due to sudden panic during fireworks display.  & 
11 deaths, 247 injured & \cite{TakashiFireworks2002} \\
\hline 
Feb 2004\Tstrut\Bstrut &  Miyun lantern festival disaster - Beijing China & Crush when a spectator stumbled on an overcrowded bridge and in the confusion people were crushed in an oncoming throng.  & 
37 deaths, 24 injured & \cite{ZhenFireworks2008} \\ 
\hline
Feb 2006\Tstrut\Bstrut &  Philsports stadium – Manilla Philippines & Sudden surged forward with tremendous speed and force when the entrance gate was flung open, coupled with steep decline and uneven surface of the road which leads to dominoes effect.  & 
74 deaths, 627 injured & \cite{LeeFire2012} \\
\hline
Nov 2008\Tstrut\Bstrut &  Wallmart black Friday shopping - New York, United States & Tension grew as the opening time for the store approaches, where the density of crowd increases rapidly and was out of control.  & 
1 death, no data is available for injured & \cite{RipleyWalmart2008} \\
\hline
Jul 2010\Tstrut\Bstrut &  Love Parade disaster - Duisburg, Germany & Crush due to unauthorized entry to the tunnel; entering fans converge with the exits. & 
21 death, 510 injured & \cite{HelbingSystemicFailure2012} \\
\hline
Nov 2010\Tstrut\Bstrut &  Khmer water festival - Phnom Penh, Cambodia & Crush caused by bottleneck on the bridge and sudden panic in crowd. & 
347 death, 755 injured & \cite{HsuCambodian2012} \\
\hline
Apr 2013\Tstrut\Bstrut &  Boston marathon bombing - Massachusetts, United States &Two pressure cooker bombs exploded near the finishing line, where the crowd of spectators gather. The suspect was later identified and found to have abandoned the bag containing the bombs nearby. & 
3 death, 264 injured & \cite{StarbirdBoston2014} \\
\hline
\end{tabular}
}
\end{table}

\end{landscape}

\clearpage

\section{Common attributes of crowd}
\label{GeneralAttributes}

In this paper, the primary focus is behavior analysis of human crowds. From the literature, it is observed that studies from the two sciences, biological and physics inspired approaches for crowd analysis in computer vision are generally based on the common attributes: decentralization, collective motion and emergent behavior. The underlying idea of crowd behavior exploited in the respective approaches is that the emergent behavior is caused by basic interactions between individuals without any force of leadership. Complex collective motions are formed, and this in turn, resulted in emergent behavior.

\subsection{Decentralized}
Decentralized decision making is the concept of having no leaders, where there is no centralized control structure to dictate how individual agents should behave. The decision arising from a process of decentralized decision making are often relate to the functional result of group intelligence and crowd wisdom in the domain of biology, or moving particles in physics. Generally, the decentralized mechanism requires positive and negative feedback, amplification and multiple interactions between agents to establish a collective unconscious that allows emergent behavior~\cite{bonabeau1999swarm}. Table \ref{Table:Decentralized} shows examples of the overlapping criteria from the perspectives of the two sciences; biology and physics.

\begin{table}[!h]
\caption{Examples of the common criteria from the perspective of a biology approach, the Ant Colony Optimization method and physics approach, the Quantum Plasma (hydrodynamic) method. }
\label{Table:Decentralized}
\centering
\resizebox{16cm}{!}{
\begin{tabular}{|p{3.5cm}|p{6cm}|p{6cm}|}
\hline 
 Criteria\Tstrut\Bstrut  & Ant Colony Optimization & Quantum Plasma\\
\hline \hline
 Positive feedback\Tstrut\Bstrut  & 
 Pheromone is left as trails that may be followed by other ants. & 
 The screening and ionic elementary excitations, at any density or temperature. \\
\hline
 Negative feedback\Tstrut\Bstrut  & 
 Pheromone slowly dissipates when the food source is exhausted. &
 The screening and ionic elementary excitations, at any density or temperature. \\
\hline
 Amplification\Tstrut\Bstrut &
 Foraging: To locate new food source. &
 Plasma oscillation in quantum plasma: `collisionless' collective mode. \\
\hline
 Multiple interactions\Tstrut &
 Ants interact with each other via trails of pheromones; successful trails are followed by more ants, thus reinforcing better routes leading to the food source. &
 Particles interactions as a coupling between density fluctuations; where the fluctuations in density produce a fluctuating internal electric field.  \\ 
\hline 

\end{tabular}
}
\end{table}

 
The decentralized behavior or decision making mechanism is indeed a distinct attribute in most of the physics and biologically inspired approaches in computer vision that give rise to emergent behavior. Simulations and observations of insect, animal and human behaviors in biology and particles in physics have supported this notion. One of the common scenarios to depict the decentralized aspect in crowd behavior is the formation of lanes (unidirectional or bidirectional) in crowded areas such as malls.

Unrelated individuals in crowd are able to create smooth traffic flows without collision by having minimal interactions according to simple rules. Individuals within a crowd use repulsive forces to reach their destination. They stay close to the shortest route between the origin and the destination, avoiding collision with obstacles or other pedestrians and rapid change of direction. 

\begin{figure}[!h]
\centering
\subfloat[Illustration of a centralized decision making, where the control is unified around a single leader (or minority decision makers).]{\includegraphics[height=0.2\linewidth, width=0.4\linewidth]{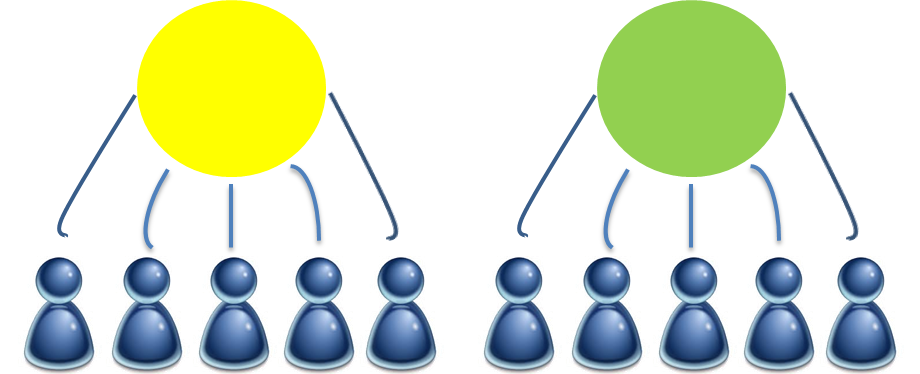}}
\qquad
\subfloat[Illustration of a decentralized decision making, where the power is with the many, not the few. There is no reliance on a particular leader to make decisions and provide directions to the swarm or group behavior.]{\includegraphics[height=0.2\linewidth, width=0.5\linewidth]{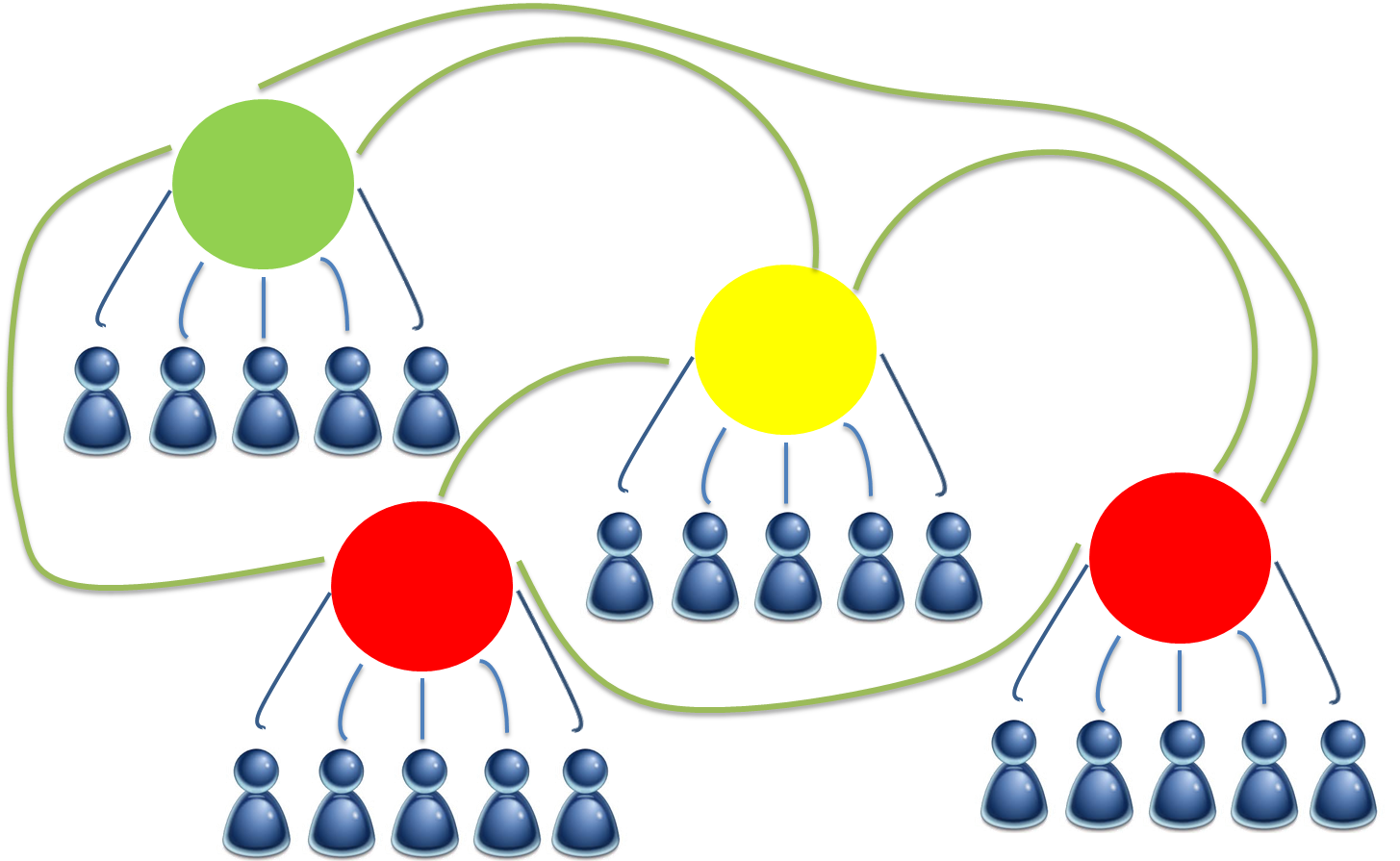}}
\qquad
\caption{Visualization of the graphical definition of centralized and decentralized decision making.}
\label{fig:Centralized}
\end{figure}

Similar models which simulate this decentralized behavior have been adapted to achieve effective evacuation planning \cite{StellaThesis2010,SabrinaSwarm2013}. 
Nonetheless, there are several researches that contradict the decentralized attributes. Crowd intelligence is disregarded by introducing leaders to include subgroup behavior \cite{Singh09,Leal11} or to allow analysis of evacuation efficiency \cite{Aube04,Pelechano06,Ji07}. Figure \ref{fig:Centralized} illustrates the graphical definition of centralized and decentralized decision making. We reckon that there is still a gap between centralize/decentralize crowd behavior model in computer vision and real world crowd scenarios. 

\subsection{Collective motion}

A comprehensive review on the observations and basic laws to describe collective motion from mathematics and physics is as presented in~\cite{Vicsek12}.  Here, the collective motion is considered as a phenomenon occurred due to ordered macroscopic behavior of constituent entities. Collective motion has been observed via simulations and experiments to not only appear in systems consisting of living beings such as human and animal~\cite{Gregoire2004,Silverberg2013,Perez2013,Miller2013}, but also amongst interacting physical objects, based on merely physical interactions without communication~\cite{Kitao1999,Erdmann2003}. Numerous models of collective patterns in humans have been suggested to describe the leading formation of such complex behaviors~\cite{Antonini2006,Burstedde2001,Willis2000,Moussaid11,Zhou2012}. Most often, what appears to make crowds unique is their ability to act in a socially coherent manner without any prior awareness, yet they are able to act as a united mass. In computer vision, a descriptor to measure the collectives of crowd and to detect collective motions is proposed in \cite{zhou2012coherent,zhou2014measuring}.

Moussaid \etal \cite{Moussaid11} proposed two heuristics that are based on cognitive information to model the desired walking directions, speeds of individuals and physical contacts between individuals in a crowd. They propose using combination of these behavioral heuristics with contact forces for a large set of complex collective dynamics.  What makes this work interesting is the proposal of integrating a cognitive science approach into the commonly physics inspired attributes for a more realistic modeling of collective social behaviors, in particular of human crowds. In another variation by Helbing and Molnar in \cite{Helbing95}, the motion of individuals is described as if they are subjected to a set of rules known as the `social force'. The `social force' is not directly exerted by the individuals' personal environment. Instead, it is a measure for the unique motivation of each individual in performing certain actions. For example, each individual motion is driven by the nearest entries or exits in a train station, and the tendency of keeping a distance or gap with other individuals to avoid collision.  This method is influenced by the Newtonian mechanics in physics. Promising results from the various studies provide evidence that the interactions between individuals in a group are indeed simple, although their resulting collective patterns are highly complex. 

\begin{figure}[!h]
\centering
\subfloat[Crowd in the pilgrimage.]{\includegraphics[height=0.21\linewidth, width=0.3\linewidth]{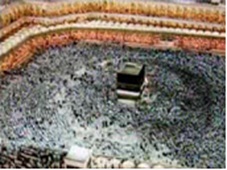}}
\qquad
\subfloat[Example of the detected abnormal region in the pilgrimage scene using the global similarity structure \protect \cite{LimICPR2014}.]{\includegraphics[height=0.21\linewidth, width=0.3\linewidth]{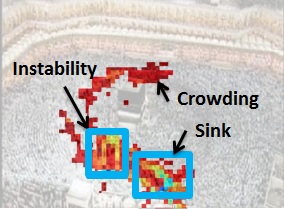}}
\qquad
\subfloat[Example of the segmented crowd region of the pilgrimage scene by treating the crowd motion as fluid \protect \cite{Ali2007}]{\includegraphics[height=0.21\linewidth, width=0.3\linewidth]{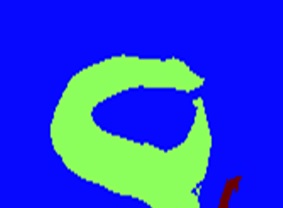}}
\caption{Examples of crowd analysis in dense crowd scenes leveraging the concept of collective motion in computer vision.}
\label{fig:Collective}
\end{figure}

Subsequently, a number of higher-level applications of crowd analysis such as detection of abnormalities within dense crowd scenes are developed by exploiting the notion of collective motion \cite{Ali2007,Solmaz2012,Loy12,LimICPR2014,cong2013abnormal,sharif2012entropy,fu2013dynamic,Noceti2014} as illustrated in Figure \ref{fig:Collective}. Their works assume that the motion of individuals tends to follow the collective flow of the crowd, and any deviations from the observed collective patterns are deemed as abnormal. Despite the promising results in estimating abnormality as well as in reproducing the observed features of collective crowd, behaviors of individuals constituting a crowd are unique in nature and thus, affect the scalability and robustness of crowd models. Their dynamics, which comprise the local and global interactions amongst the group and environment further complicate analysis. A summary of the understanding of collective motion from the perspectives of biology and physics is described in Table \ref{Table:CollectiveMotion}.

\begin{table}[!h]
	\caption{Detailed description of the understanding of the common criteria in collective motion for the two models; social force model \cite{Helbing95} and cognitive heuristics model \cite{Moussaid11}. }
	\label{Table:CollectiveMotion}
	\centering
	\resizebox{16.5cm}{!}{
		\begin{tabular}{|p{2cm}|p{9cm}|p{9cm}|}
			\hline 
			Criteria\Tstrut\Bstrut  & Cognitive Heuristics Model \cite{Moussaid11} & Social Force Model \cite{Helbing95}\\
			\hline \hline
			Concept or formulation\Tstrut\Bstrut  & 
			A cognitive science approach, based on behavioral heuristics (biology). & 
			A force-based model inspired by Newton Dynamics (physics). \\
			\hline
			Negative feedback\Tstrut\Bstrut  & 
			Pedestrian behavior is guided by visual information (within line of sight) by assuming that:
			\begin{itemize} [leftmargin=+0.1in]
				\item Pedestrian seeks an unobstructed walking direction, without deviating too much from the direct path to destination.
				\item The desired walking speed is influenced by the need to maintain a distance from the obstacle in the walking direction that ensures a safe interval to avoid collision (intentional displacement).
				\item The interaction forces between bodies during overcrowding (unintentional displacement).
			\end{itemize}
				
				&
			Pedestrian behavior can be modeled as equation of motion determined by:
			\begin{itemize} [leftmargin=+0.1in]
				\item The goal to reach destination as comfortable as possible.
				\item Repulsive effect: the need to remain a comfortable distance with the surroundings.
				\item Attractive effect: the tendency to form groups.
			\end{itemize}
				
			\\
			\hline
			Interaction\Tstrut\Bstrut &
			The interaction terms are non-zero only in extremely crowded situations, and not under normal walking conditions. &
			Pedestrian walking behavior is influenced by the repulsive and attractive force with its surrounding. \\
			\hline
			Application \Tstrut &
			Example Scenarios:
			\begin{itemize} [leftmargin=+0.1in]
				\item The avoidance of obstruction or other pedestrian under both, the unidirectional and bidirectional flows. 
				\item The self-formation of lanes consisting of pedestrians with a uniform walking direction, under varying density level (low to high; smooth flows to stop-and-go waves and crowd turbulence).
				\item Crowd turbulence in panic situations caused by unintentional collisions and bottleneck. 
			\end{itemize}
			&		
			Example Scenarios:
			\begin{itemize} [leftmargin=+0.1in]
				\item The self-formation of lanes consisting of pedestrians with a uniform walking direction
				\item Oscillatory changes of the walking direction at narrow passages. 
			\end{itemize}  
			\\ 
			\hline 
			
		\end{tabular}
	}
\end{table} 

\subsection{Emergent behavior}

\begin{figure}[!h]
\centering
\includegraphics[height=0.41\linewidth, width=0.98\linewidth]{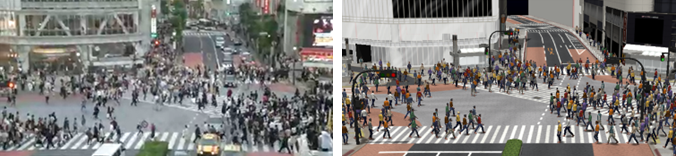}
\caption{Comparison between the actual scene (right) and the simulated scene (left) of Shibuya Crossing in Japan.  The proposed least-energy model by \protect Guy \etal \cite{GuyEmergent2012} is able to reproduce the emergent behavior in actual crowd scenes, such as arching and self-organization into lanes, by applying simple and intuitive formulation of rules and biomechanical measurements to the individuals agents and their interactions.}
\label{fig:Shibuya}
\end{figure}

Emergence is the process of complex pattern formation from simple rules. Thus an emergent behavior or emergent property can arise when a number of individuals operate collectively in an environment, forming complex behaviors. A critical distinction between the emergent behavior and collective motion is that the former is a \textbf{phenomenon resulting from collective motion}, while the later describes the \textbf{self-organization pattern of crowd}. In both nature and engineering, complex behaviors can emerge as a result of distributed collective processes or collective group behavior.  The physicists, biologists, philosophers and computational scientists have for a long time study the emergent properties in their respective domains to shed light into the problem of understanding emergent behavior \cite{Kitto2006}. From the physics point of view, the emergence of complex behaviors has been observed in photons and electrons in quantum systems and fluid dynamics \cite{Dutton75,Swenson89,Laughlin2000,Riccardo2006,Cubrovic09}. Here, the collective phenomenon observed in macroscopic systems is brought about by the microscopic constituents with one another and with their environment. A biological example of the emergent behavior is an ant colony. An ant as a single entity has limited memory and is capable to perform only simple actions. However, an ant colony expresses a complex, collective behavior which provides intelligent solutions to problems such as finding the shortest route from the nest to a food source \cite{Blum2005,Ratnieks2008,Dorigo08,Dorigo2003}. Figure \ref{fig:Shibuya} illustrates an example scenario of the emergent behavior in human crowd. Meanwhile, Table \ref{Table:EmergentBehaviour} describes examples of emergent behavior in physics and biology. 

\begin{figure}[!h]
\centering
\includegraphics[height=0.21\linewidth, width=1\linewidth]{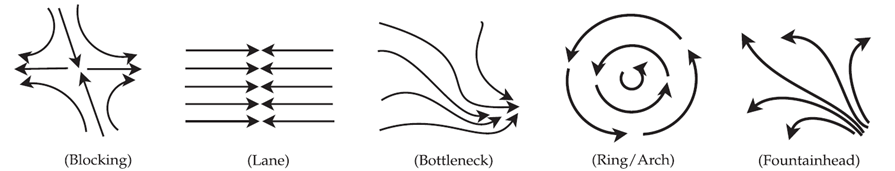}
\caption{Five crowd behavior patterns identified in \protect \cite{Solmaz2012}. }
\label{fig:Crowdpatterns1}
\vspace{5mm}
\includegraphics[height=0.41\linewidth, width=0.94\linewidth]{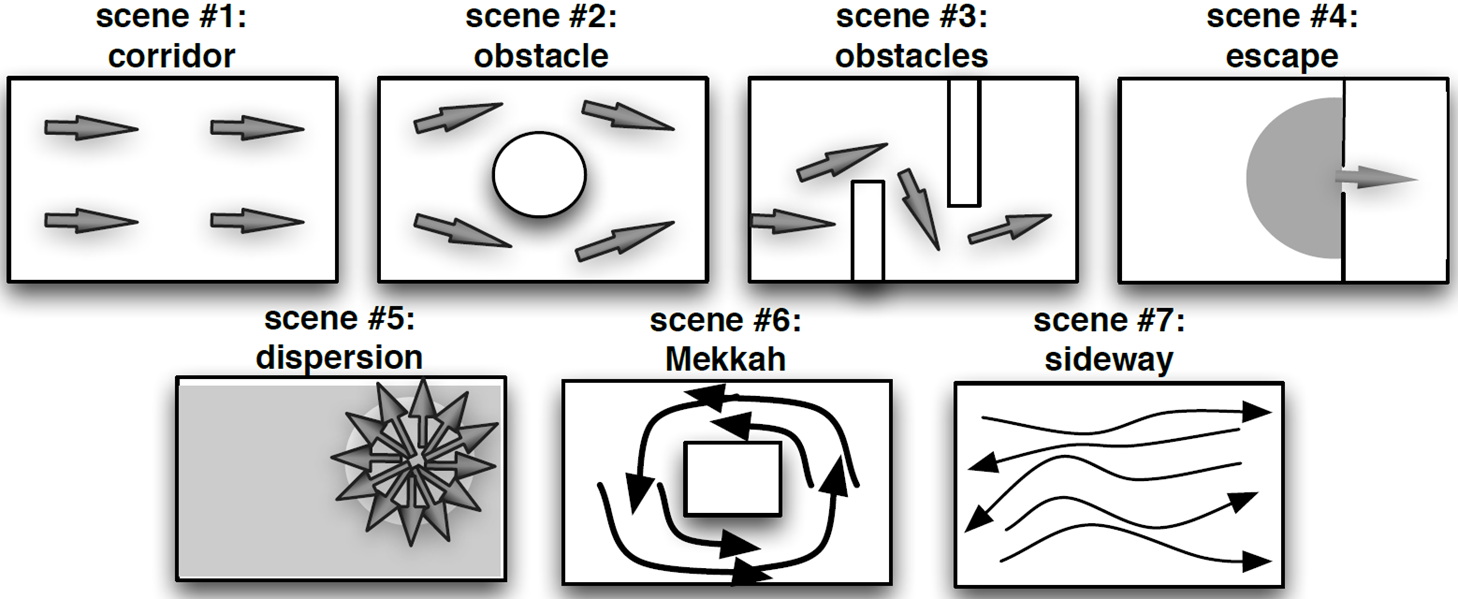}
\caption{The different crowd behavior scenes proposed in \protect \cite{Allain2012}.}
\label{fig:Crowdpatterns2}
\end{figure}

The underlying theory that we observe to be consistent across the wide range of domains is that, the emergent of the \textbf{highly complex behavior} in a collection of individuals (i.e. animal, electron, neuron, particle) is often a result of individuals following a set of \textbf{simple rules}. Furthermore, these interactions are established to be \textbf{decentralized}, without any external coordination or the active role of a leader. Another characteristic that is consistent across experts from different areas is that emergent behavior is often \textbf{unpredictable and unprecedented}. Work that focus on the emergent behavior of crowd includes ~\cite{Solmaz2012}, where a linear approximation of the dynamic system is used to categorize varying emergent crowd behaviors based on eigenvalues over an interval of time. Their method classifies crowd behaviors into bottleneck, lane, arch, fountainhead and blocking (as illustrated in Figure~\ref{fig:Crowdpatterns1}). Likewise, Allain \etal \cite{Allain2012} focuses on seven key crowd behaviors generated using  Lagrangian forces as shown in Figure~\ref{fig:Crowdpatterns2}. Despite the promising results, questions may arise on the adequacy on imposing selected predefined `patterns' and set of `rules' in existing models to capture the complexity of real-world scenarios. 

\begin{table}[!h]
	\caption{Examples of emergent behaviors in biology and physics.  }
	\label{Table:EmergentBehaviour}
	\centering
	\resizebox{16cm}{!}{
		\begin{tabular}{|p{2.5cm}|p{6.5cm}|p{6.5cm}|}
			\hline 
			Criteria\Tstrut\Bstrut  & Emergent Behaviour in Human Crowd (Biology)
			& Emergent Behaviour in Crystalline Solids (Physics)
			\\
			\hline \hline
			Entity\Tstrut\Bstrut  & 
			Human Crowd & 
			Particle \\
			\hline
			Decentralized decision making \Tstrut\Bstrut  & 
			Each individual in the crowd follow simple rules (i.e. towards destination, avoid collision), without any centralized control or communication to dictate how an individual should behave, or act. 
			&
			Particle interactions as a coupling between density fluctuations, without any centralized or fixed control to dictate the type of interaction. 
			\\
			\hline
			Collective motion or behavior\Tstrut\Bstrut &
			Interact locally with one another and with the environment (i.e. to avoid collision with others or walls). &
			The plasma oscillation is an example of a `collisionless' collective mode, in which the restoring force is an effective field brought about by particle interaction. \\
			\hline
			Emergent behavior \Tstrut &
			Crowd dynamics - In a bidirectional traffic in a street where individuals move at random positions, it can be observed that the flow directions separate spontaneously after a short time (lane formation phenomena).
			&		
			Crystalline solids - At high enough temperature, any form of quantum electronic matters becomes a plasma. Then as it cools down, a plasma will become liquid and as the temperature falls further, it turns into a crystalline solid.  
			\\ 
			\hline 
			
		\end{tabular}
	}
\end{table} 

\section{Conflicting attributes of crowd}
\label{ContradictingAttributes}

In the research of crowd behavior analysis, abstractions are made regarding motions of individuals in crowd. The two main aspects discussed in this section is the tendency and capability of individuals to think and have preferences in the crowd. 

\subsection{Bias or non-bias}

Biologically, entities in crowd have natural tendency or predilection. A fairly good example is the schooling species of golden shiners (\textit{Notemigonus crysoleucas}) where they have pre-existing bias towards yellow targets. Similarly, every individual is exclusive where each has their own unique behavior and inclination. Individuals in crowd are `bonded' by one common focal points \cite{raey2005}, in which each individual has low relatedness with varying self-interest \cite{couzin2011uninformed}. Hence, the task to understand and model the varying predilection and behavior of each entity is complex. 

Nevertheless, when individuals are integrated as a crowd, they unconsciously alter their behavior in line with the responses of neighboring entities. As exhibited in \cite{couzin2011uninformed}, despite the presence of powerful minority with strong inclination, with sufficient amount of uninformed entities, the crowd will come to a majority decision. In another work by Klucharev \etal \cite{klucharev2009reinforcement}, they found that when individuals are made aware of the `opinions' of the crowd, individuals will adjust their judgments to align with the opinions of the general crowd. This behavior is in fact referring to the natural reflect that is deeply rooted in each entity (specifically human) to conform to social norm; which comply with principles of reinforcement learning \cite{klucharev2009reinforcement,haun2012majority}. When there is disparity with the general norm of crowd, neuronal responses are triggered and manifested in the rostral cingulated zone and ventral striatum in the brain \cite{klucharev2009reinforcement}, which leads to tendency to adapt.

The natural responses of a human crowd to conform to social norm and the general crowd makes it tolerable to simplify and relax the `biasness' at individual level when modeling and understanding crowd behavior in both biological and physic inspired approaches. Crowd behavior models emphasize more on the collective effect of entities - emergent behavior \cite{henein2010microscopic}. In \cite{Ali2008}, an algorithm to track the path maneuvered by individuals in a dense crowd based on the observation that locomotive pattern of an individual is in line with the collective patterns of the crowd and the layout of the environment is proposed. Alternatively, in \cite{Zhou2012}, the idea of conformity of individual to crowd is used to infer the past and future behavior of individual. Then again, to analyze the crowd behaviors in a real-world scenario, questions might arise as up to which extent an individual is conforming to the norm of general crowd.

\subsection{Thinking or non-thinking}
Originally, works on crowd behavior are frequently associated with the laws of physics, presuming that it is sensible to view crowd as a homogeneous mass of bodies \cite{sime1995crowd,low2000statistical}. Each individual within the crowd is characterized as a non-thinking particle which has no tendency and capability to make decisions. The motions and directions of each particle are dictated by the external forces (i.e. boundaries, neighboring particles, etc.) \cite{Moore11}. For instance, motion of mass of individual is equated with the flow of fluid steered by the pathway with the rate of motions computed analogue with the law of fluid dynamic.

In 1995, Sime \cite{sime1995crowd} questioned the practicality of representing human in crowd as non-thinking particles (or ball bearings); forsaking the rules of behaviors. Often, the minute behaviors and reactions of individuals within the crowd to surrounding are the vital interactions that affects the crowd motion as a whole. For example, a person who fell can become an obstacle to a smooth flow of a crowd, that may deteriorate and lead to the occurrence of stampede and death. The lack of behavioral complexity in crowd models makes it an imprecise description of `real life' crowd flow. With that being said, many new approaches integrated physic and psychological aspects in crowd behavior analysis; exploring crowd behavior from both macroscopic (crowd as a whole) and microscopic (interactions between individuals) perspective. Among the new approaches is the model proposed by Helbing \etal \cite{HelbingHerding2000}. The authors modeled crowd behavior by simulating the tendency of individuals to conform to social norm and reactions (i.e. panic) evoked as response to the layout and atmosphere of the environment. 

\begin{figure}[!h]
\centering
\includegraphics[height=0.48\linewidth, width=0.8\linewidth]{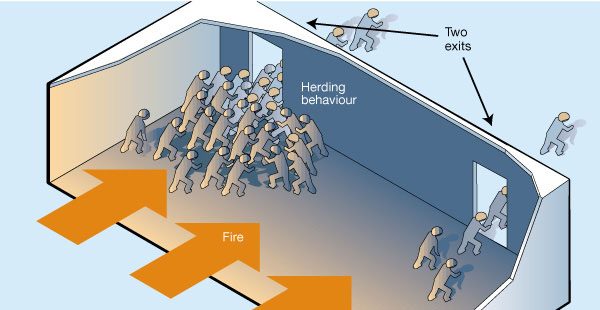}
\caption{Most simulations of pedestrian behavior in crowds, especially in modeling evacuation, used models based on fluid flow through pipes, which ignores the actions or `thinking' component of the individuals. An example work by Helbing \etal in \protect \cite{HelbingHerding2000} shows that the evacuation of pedestrians from a smoke-filled room with two exits may lead to herding behavior and clogging at one of the exits. By contrast, a traditional fluid-flow model would have predicted an efficient use of both of the exits. The contradicting finding by Helbing \etal which matches more closely with the actual scenario during evacuation is made possible by introducing a mixture between the individualistic and collective behavior. Their findings provide preliminary results that are worthy of future investigations in order to understand the `thinking' and `non-thinking' components as well as the `bias' and `non-bias' aspects of crowd; with the aim to provide better solutions that are able to mitigate the negative impact of the emergent behaviors such as herding and clogging, towards creating a safer environment. }
\label{fig:Herding}
\end{figure}

Nevertheless, computational complexity is directly proportional to the density of a crowd. There is a dire need to achieve equilibrium between the understanding that individuals within a crowd are capable of making decision and the incorporation of the amount of influence an individual behavior has to the others in the crowd. Figure \ref{fig:Herding} demonstrates an example scenario where the behaviors of crowd in the real-world differ from the common assumption of physics inspired models. This stir thoughts and interest on the need to understand further the varying perspectives on the attributes of crowd (bias or non-bias, thinking or non-thinking).   

\section{Applications \& Dataset}
\label{Applications}

Crowd behavior analysis and understanding is a subject of great scientific interest that involve multidisciplinary field, including computer vision, psychology, sociology, physics and civil engineering to understand the multifaceted of the study of crowd. Despite the possibility of being an inexhaustible source of research, it is motivated by the need for an enhanced public safety in the society~\cite{Lim2014}.

In the following section, we will discuss the public datasets corresponding to the three branches of applications of crowd behavior analysis: i) crowd segmentation, ii) crowd dynamic analysis, iii) crowd density estimation. In particular, the discussion is driven by the methods to model the collective motion and the contextual understanding of the emergent behavior for various applications. 

\subsection{Dataset}

As crowd behavior analysis in the field of computer vision prosper, public datasets start to gain importance in the vision community to meet different research challenges. Please refer to Table \ref{Table:Dataset} and Table \ref{Table:Dataset2} for a complete list of the currently available datasets.

Although different researches from varying scientific fields are trying to analyze the same physical entity (i.e. crowds of individuals), the approaches have advanced independently \cite{ali2013modeling}. As such, different techniques, comparative comparisons and benchmark datasets developed are characteristically of its own; resulting in the difficulty to summarize the evaluation protocol and performance comparisons in this area. In an effort to make the next great leap forward in crowd behavior analysis, we strongly believe that there is a need of a common platform to evaluate the various analytical aspects of crowd. 

\clearpage

\begin{landscape}
\begin{table} 
\centering
\caption{Publicly available datasets for crowd behavior analysis. }
\label{Table:Dataset}
\resizebox{22cm}{!}{
\begin{tabular}{|m{1.7cm}|c|m{4.5cm}|m{20cm}|}
 \hline
\Tstrut Established Year\Bstrut & Dataset Reference & Dataset Name & Details  \\
\hline \hline


\hline
2007\Tstrut\Bstrut  & \cite{Ali2007} & UCF dataset - crowd segmentation & Crowd videos collected mainly from BBC Motion Gallery and Getty Images website comprising scenes of human crowd and other high density moving objects. Crowd density in scenes ranges from medium to extremely dense with variation in perspectives and illumination. \\
 
\hline
2008\Tstrut\Bstrut  & \cite{Ali2008} & UCF dataset - crowd tracking & Contain three marathon sequences with its corresponding static floor fields.   \\ 


\hline
2008\Tstrut\Bstrut & \cite{chan2008privacy} & UCSD Pedestrian Dataset  & Video sequences are recorded at USCD campus walkways using hand-held camera from two viewpoints. All videos are 8-bit gray scale, at 10fps with resolution of 238x158. One of the scenes contains sparse pedestrian traffic, another one with large crowd moving up the walkway.  \\

\hline
2008\Tstrut\Bstrut & \cite{loy2008local} & QMUL Junction dataset  & A one hour long video sequence of busy traffic captured at 25fps with resolution of 360x288. The traffic is regulated by traffic lights and dominated by four types of traffic flows.    \\

\hline
2009\Tstrut, 2010 \& 2012\Bstrut & \cite{Ferryman2009,Ferryman2010} & PETS 2009 / Winter-PETS 2009  / PETS 2010 / PETS 2012 
& This dataset is a collection of videos obtained from multiple sensors, which was introduced by the Performance Evaluation of Tracking and Surveillance Workshop (PETS) since year 2000. Over the years, the dataset presents a broader scope of scene understanding challenge, from low-level video analysis such as object tracking to mid-level analysis such as people falling, and lastly, to high-level analysis such as threat event detection. The datasets highlighted here are the ones commonly used for crowd analysis.\\

\hline
2009\Tstrut\Bstrut & \cite{umndataset2009} & UMN Crowd dataset & Consists of 11 difference scenarios of an escape event in three different indoor and outdoor scenes. All sequences starts with normal behavior followed by sequences of abnormal behavior.  \\

\hline
2010\Tstrut\Bstrut & \cite{Mahadevan2010} & UCSD Anomaly Detection dataset & A set of 100 video sequences acquired with a stationary camera mounted at an elevation, overlooking pedestrian walkway. Density of crowd varies from sparse to very crowded. Anomalies in the scene are due to: i) circulation of non-pedestrian entities in the walkway, or ii) anomalous pedestrian motion patterns. The ground truth annotation includes a binary flag per image frame indicating presents of anomaly and manually generated pixel-level binary masks (to localize regions of anomalies).  \\

\hline
2011\Tstrut\Bstrut & \cite{Raghavendra2011} & Crowd and group data 
& Comprises 38 video sequences, which combines existing datasets (UMN, PETS 2009 and UCF) and additional real scenarios videos obtained from Youtube.com and ThoughtEquity.com. The collection of videos covers a wide selection of the different scenarios of anomaly in crowd, ranging from sparse, to medium and high density crowd. Each video starts with the normal behavior where the crowd motion is regular, followed by a sequence of abnormal behavioral frames such as sudden dispersal of the crowd, crowd running towards random directions and high interactions between individuals in the crowd caused by brawling.   \\

\hline
2011\Tstrut\Bstrut & \cite{Rodriguez11} & Data-Driven Crowd dataset & Contain unique real crowd videos (11GB in total) collected by crawling and downloading from search engines and stock footage websites (such as Getty Images and YouTube). Each video ranges between two to five minutes with resolution of 720x480. The dataset does not include time-lapse videos and videos taken from tilt-shift lenses.  $\;\;\;\;\;$ \hspace{50pt} \\

\hline
2012\Tstrut\Bstrut & \cite{Zhou2012} & Train Station dataset & A 33.2 minutes long video sequence collected from the New York Grand Central Station. The video sequence is 25fps with a resolution of 720x480. The corresponding KLT key point trajectories extracted from the video is provided. \\

\hline
2012\Tstrut\Bstrut & \cite{Chen2012} & Mall dataset & Public surveillance footage in a shopping mall with challenging lighting and reflective glass surface. The density of crowd ranges from sparse to crowded with varying behavior (stationary and dynamic crowds). Ground truth consists of annotation of over 60,000 pedestrians (labels on head position) in 2000 video frames. Resolution of each frames is 640x480.  \\ 

\hline
2012\Tstrut\Bstrut & \cite{Allain2012} & AGORASET & Simulation-based crowd video dataset composed of eight scenes generated using simulation model based on Lagrangian forces by Helbing \etal \cite{HelbingHerding2000}. The videos correspond to various conditions (i.e. illumination, viewing angle, stress level of the crowd, etc.). The associated ground truth provided is in terms of individual trajectories and related continuous quantities of each scene (i.e. density and velocity field). \\ 

\hline
2012\Tstrut\Bstrut & \cite{hassner2012violent} & Violent-Flows dataset & A database collected from YouTube consists of 246 real crowd violence and non-violence video footage. The duration of each video ranges between 1.04 and 6.52 seconds. The footages are of different types of scenes in uncontrolled condition with varying video qualities and surveillance scenarios. $\;\;\;\;\;$ \hspace{100pt} \\ 

\hline
2013\Tstrut\Bstrut & \cite{Haroon2013} & UCF dataset - crowd counting & Consists of 50 crowd images collected mainly from FLICKR with ground truth annotation. The counts range between 94 and 4543 individuals per image. Crowd scenes belong to diverse events: concerts, protests, stadiums, marathons and pilgrimages. \\ 

\hline
2014\Tstrut\Bstrut & \cite{shao2014scene} & CUHK Crowd dataset & A dataset of 474 video sequences from 215 crowded scenes collected from Pond5, Getty-Images and manually captured by the authors. It includes scenes with various densities and perspective scales captured from different environment. The associated trajectories extracted using GKLT tracker \cite{zhou2014measuring} for each video are provided. \\ 

\hline
2014\Tstrut\Bstrut & \cite{LimICPR2014} & Crowd Saliency dataset & Comprises 20 videos obtained from various sources, such as the UCF and Data-driven crowd datasets. The sequences are diverse, representing dense crowd in the public spaces in various scenarios such as pilgrimage, station, marathon, rallies and stadium. In addition, the sequences have different fields of view, resolutions, and exhibit a multitude of motion behaviors that cover both the obvious and subtle instabilities for saliency detection. The ground truths of salient region in each video are provided.  \\ 

\hline
\end{tabular}
}
\end{table}

\end{landscape}

\clearpage

\begin{landscape}
	\begin{table} 
		\centering
		\caption{Publicly available datasets introduced for specific crowd behavior analysis with the reported quantitative results from the corresponding reference.}
		\label{Table:Dataset2}
		\resizebox{21cm}{!}{
			\begin{tabular}{m{6cm}cc|c|c|c}
				\hline
				\Tstrut Dataset Name\Bstrut & Established Year & Dataset Reference & Crowd Segmentation & Crowd Dynamic Analysis & Crowd Density Estimation \\
				\hline \hline
				
				UCF dataset - crowd segmentation & 2007\Tstrut\Bstrut  & \cite{Ali2007} &  (Qualitative evaluation) & (Qualitative evaluation) & - \\
				
				\hline
				UCF dataset - crowd tracking & 2008\Tstrut\Bstrut  & \cite{Ali2008} & \begin{minipage}[t]{0.52\textwidth} \begin{itemize}[noitemsep]\item [] \underline{Person tracking accuracy} \item [] Marathon-1: 74.9\% (143/199 individuals) \item [] Marathon-2: 97.5\% (117/120 individuals) \item [] Marathon-3: 76\% (38/50 individuals) \vspace{0.52em}\end{itemize} \end{minipage} & - & -\\ 
				
				\hline
				UCSD Pedestrian Dataset  & 2008\Tstrut\Bstrut & \cite{chan2008privacy} & \begin{minipage}[tb]{0.52\textwidth} \begin{itemize}[noitemsep] \item [] \vspace{0.5em} \underline{Motion segmentation} \item [] True positive rate: 0.936 \item [] False positive rate: 0.036 \item [] Area under ROC: 0.9727 \vspace{0.5em} \end{itemize} \end{minipage}  & - & \begin{minipage}[tb]{0.52\textwidth} \begin{itemize}[noitemsep] \item [] \vspace{0.5em} \underline{Mean squared error} \item [] Away: 4.181 \item [] Towards: 1.291 \vspace{0.5em} \item [] \underline{Absolute error} \item [] Away: 1.621 \item [] Towards: 0.869 \vspace{0.5em} \end{itemize} \end{minipage} \\
				
				\hline
				QMUL Junction dataset  & 2008\Tstrut\Bstrut & \cite{loy2008local} & - & \begin{minipage}[tb]{0.5\textwidth} \begin{itemize}[noitemsep] \item [] \vspace{0.5em} \underline{Anomaly detection} \item [] Area under ROC: 0.9765 \vspace{0.5em} \end{itemize} \end{minipage} & - \\
				
				\hline
				PETS 2009 / Winter-PETS 2009  / PETS 2010 / PETS 2012 & 2009\Tstrut, 2010 \& 2012\Bstrut & \cite{Ferryman2009,Ferryman2010} & - & - & - \\
				
				\hline
				UMN Crowd dataset & 2009\Tstrut\Bstrut & \cite{umndataset2009} & - & - & - \\
				
				\hline
				UCSD Anomaly Detection dataset & 2010\Tstrut\Bstrut & \cite{Mahadevan2010} & - & \begin{minipage}[t]{0.5\textwidth} \begin{itemize}[noitemsep] \item [] \underline{Anomaly detection} \item [] Equal error rate: 25\% \vspace{0.5em} \item [] \underline{Anomaly Localization} \item [] Rate of detection: 45\% \vspace{0.5em}  \end{itemize} \end{minipage} & - \\
				
				\hline
				Crowd and group data & 2011\Tstrut\Bstrut & \cite{Raghavendra2011} & - & \begin{minipage}[t]{0.5\textwidth} \begin{itemize}[noitemsep] \item [] \underline{Global anomaly detection} (Area under ROC) \item [] UMN dataset: 0.9961 \item [] Prison riot dataset: 0.8903 \item [] UCF dataset: 0.986 \item [] PETS 2009: 0.9414 (scene 1), 0.9914 (scene 2) \vspace{0.5em} \end{itemize} \end{minipage} & - \\
				
				\hline
				Data-Driven Crowd dataset & 2011\Tstrut\Bstrut & \cite{Rodriguez11} & \begin{minipage}[t]{0.5\textwidth} \begin{itemize}[noitemsep] \item [] \underline{Tracking typical crowd behavior} \item [] Mean tracking error: 47.47$\pm$1.27 pixels \vspace{0.5em} \item [] \underline{Tracking rare/abrupt events} \item [] Mean tracking error: 46.88 pixels \vspace{0.5em} \end{itemize} \end{minipage} & - & - \\
				
				\hline
				Train Station dataset & 2012\Tstrut\Bstrut & \cite{Zhou2012} & - & (Qualitative evaluation) & -\\
				
				\hline
				Mall dataset & 2012\Tstrut\Bstrut & \cite{Chen2012} & - & - & \begin{minipage}[t]{0.52\textwidth} \begin{itemize}[noitemsep]  \item [] Mean absolute error: 3.15 \item [] Mean squared error: 15.7 \item [] Mean deviation error: 0.0986 \vspace{0.5em}  \end{itemize} \end{minipage}\\ 
				
				\hline
				AGORASET & 2012\Tstrut\Bstrut & \cite{Allain2012} & - & - & - \\ 
				
				\hline
				Violent-Flows dataset & 2012\Tstrut\Bstrut & \cite{hassner2012violent} & - & \begin{minipage}[t]{0.52\textwidth} \begin{itemize}[noitemsep] \item [] Crowd violence video classification: 81.3$\pm$0.21\% \item [] Crowd violence detection: 88.23\% \vspace{0.5em} \end{itemize} \end{minipage} & - \\ 
				
				\hline
				UCF dataset - crowd counting & 2013\Tstrut\Bstrut & \cite{Haroon2013} & - & - & \begin{minipage}[t]{0.52\textwidth} \begin{itemize}[noitemsep] \item [] Absolute Difference/image: 419.5$\pm$541.6 \item [] Normalized Absolute Difference/image: 31.3$\pm$27.1 \vspace{0.5em} \end{itemize} \end{minipage} \\
				
				\hline
				CUHK Crowd dataset & 2014\Tstrut\Bstrut & \cite{shao2014scene} & \begin{minipage}[t]{0.52\textwidth} \begin{itemize}[noitemsep] \item [] \underline{Group detection} \item [] Normalized mutual information: 0.48 \item [] Purity: 0.78 \item [] Rand index: 0.83 \vspace{0.5em} \end{itemize} \end{minipage} & \begin{minipage}[t]{0.52\textwidth} \begin{itemize}[noitemsep]\item [] \underline{Group state analysis} \item [] Average accuracy: 60\% \vspace{0.5em} \item [] \underline{Crowd video classification} \item [] Average accuracy: 70\% \vspace{0.5em} \end{itemize} \end{minipage} & -\\ 
				
				\hline
				Crowd Saliency dataset & 2014\Tstrut\Bstrut & \cite{LimICPR2014} & - & \begin{minipage}[t]{0.52\textwidth} \begin{itemize}[noitemsep] \item [] \underline{Crowd salienct detection} \item [] (No. of detection/Labeled region) \item [] Crowding: 12/13 (1 missed detection) \item [] Sources \& sinks: 14/19 (5 missed detection) \item [] Local irregularity: 47/43 (2 missed detection, 6 false detection) \vspace{0.5em}\end{itemize} \end{minipage}  & - \\ 
				
				\hline
			\end{tabular}
		}
	\end{table}
	
\end{landscape}

\subsection{Crowd segmentation}
Generally, works in crowd segmentation assume that crowd is an agglomeration of pedestrians \cite{Zhou2012}. Even though each individual has their own goal destination and motion tendency, they appear to share common motion dynamics when observed over time in a crowded scene. This is due in part to the tendency of individuals to follow the dominant flow owing to the physical structure of the scene, and the social conventions of the crowd dynamics \cite{LimICPR2014}. Therefore, crowd segmentation is commonly the basis of many abnormal event detection systems, whereby finding the abnormal regions in a given scene is accomplished by discovering the deviation from the regular flow, also known as coherent motion, as stored in the crowd motion model. In some scenarios, crowd segmentation is applied prior to estimating the density of crowd \cite{ZhangSegmentation2012}.  

\begin{figure}[!h]
\centering
\subfloat[Input sequence: Crowd in the pilgrimage or Haji scene.]{\includegraphics[height=0.32\linewidth, width=0.45\linewidth]{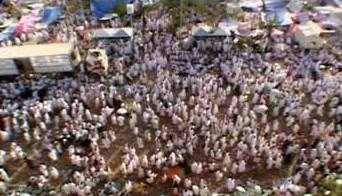}}
\qquad
\subfloat[Example of the segmented regions, where regions with similar behavior in the Lyapunov sense are merged as proposed in \protect \cite{Ali2007}. The different colors in the results represent different flow segments.]{\includegraphics[height=0.32\linewidth, width=0.45\linewidth]{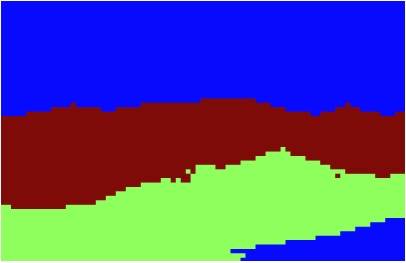}}
\qquad
\subfloat[Input sequence: Crowd in the marathon scene.]{\includegraphics[height=0.32\linewidth, width=0.45\linewidth]{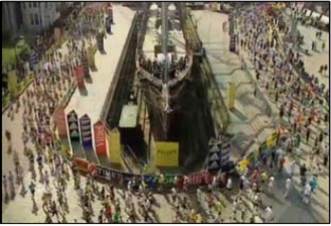}}
\qquad
\subfloat[Example of the segmented regions, where the region growing scheme is used to segment crowd flow based on its flow field as proposed in \protect \cite{WuSegmentation2009}.]{\includegraphics[height=0.32\linewidth, width=0.45\linewidth]{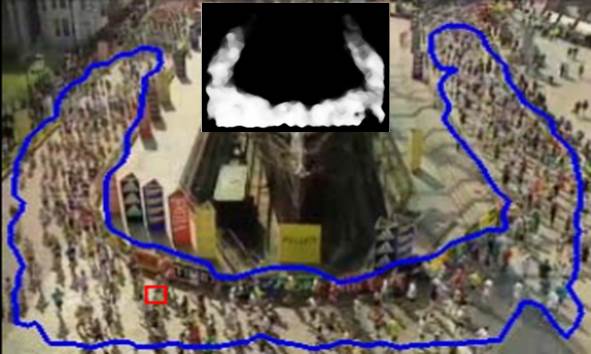}}
\caption{Examples of the crowd segmentation results from state-of-the-art methods.}
\label{fig:Segmentation}
\end{figure}

The main focus of crowd segmentation methods is on grouping regions with similar motion dynamics or coherency \cite{wu2012joint}. Most often, rather than computing the trajectories of individuals (microscopic), holistic approaches (macroscopic) in crowd segmentation methods build a crowd motion model using instantaneous motions of the entire scene such as the flow field \cite{Ali2007,mehran2010streakline,WuSegmentation2009}. There are, however some work which is based on tracking individuals and accumulating their trajectories over a period of time to obtain coherent motion \cite{Zhou2012}. Tracking approaches, regardless of whether they are using distance or model-based representations are very challenging in crowd scenes \cite{WangTracklets2013}. This is because the trajectories are highly fragmented with many missing observations due to the complex interactions, occlusions between individuals in the crowd and background clutters. Therefore, tracking in crowded scenes often incorporate scene or contextual information for accurate trajectory estimation \cite{DehghanTracking2012}. Although coherent motion is generally the macroscopic and microscopic observations of collective movements of individuals, recent studies show that it can also be characterized by mesoscopic models. Mesoscopic approaches such as in~\cite{Zhou2012,MoussaidSwarm2009,WangTracklets2013} model crowd activities using the interactions between an individual and its local neighborhood.  In \cite{ZhouTopic2011}, fragments of trajectories over a short period of time, known as tracklets are used to analyze motion coherency.  The general view of crowd is nevertheless retained, by incorporating the information on the sources and sinks. Their model encourages the tracklets to have the same sources and sinks as the segmented regions. Despite the fact that similarly crowd detection algorithm proposed by Reisman \etal \cite{reisman2004crowd} works in the spatio-temporal domain, the system relies on the inward and intersection motions of opposite moving individuals to infer presence of crowd.  Examples of the segmentation results from related works are as illustrated in Figure \ref{fig:Segmentation}.

\begin{figure}[!h]
\centering
\includegraphics[height=0.3\linewidth, width=1\linewidth]{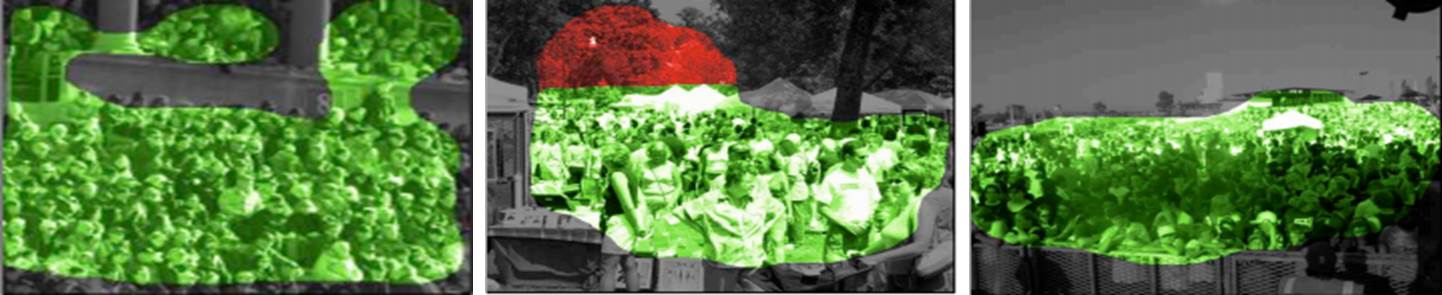}
\caption{Sample results of crowd segmentation from another perspective, where regions containing crowd are segmented accordingly as proposed in \protect \cite{Arandjelovic2008}. The true positives are highlighted in green whereas the false positives are represented by the red areas.}
\label{fig:Detection}
\end{figure}

While the earlier discussed works are fixated on segmenting coherent motions as a cue of crowd on videos or image sequences, another variant perform crowd detection on still images as proposed in \cite{Arandjelovic2008}. Their work allows discrimination between crowd and non-crowd regions by utilizing low-level local feature from single crowd image. Responses for each pixel are defined using pyramid pixel-grid approach to exploit the properties of crowd (at narrow scale, the basic element should resemble a human; whereas at large scale, crowd regions exhibit repetitive features of individuals in crowd). Sample results are shown in Figure \ref{fig:Detection}. Similarly, Fagette et al.~\cite{Fagette2014} proposed an unsupervised method using multi-scale texture-based features to detect and localize dense crowds in images. The unsupervised method allows detection on images without the need to have prior knowledge of the scene or context. In another variation, Idrees et al.~\cite{Idrees2015} proposed a new direction to localize crowd segments by detecting humans in dense crowds. Their detection method uses the collective crowd attribute based on the observation that scale of individuals in local neighborhood is similar. 

Regardless of the low-level features used to segment regions of crowd, crowd segmentation methods are indirectly based on the assumptions that there is indeed collective motion between individuals in the crowd. The collective motion between individuals that move together with consistent speed and motion direction describes the self-organization pattern of crowd and can be observed commonly in public spaces such as mall and underground station. In crowd segmentation, the phenomenon arising from the collective homogeneous and coherent motion such as the formation of unidirectional lanes and shortest path is the emergent behavior. In real world applications, the emergent behavior changes according to time (e.g. traffic flow and crowd density increases during peak hours) and thus, segmentation methods must be flexible enough to cope with the differing motion activity. 

\subsection{Crowd dynamic analysis}

The formation of crowd and mass gathering often poses challenges to public safety if it is not handled effectively, particularly when panic arises among surging individuals \cite{helbing2002simulation}. Therefore, amongst the major goal of computer vision systems is to detect and analyze the motion dynamics of crowded scenes, in the hope towards profiling and identifying salient motion behaviors which could lead to potential unfavorable events. In the literature, there has been a variety of terms used to refer to salient motion behavior including interesting, irregular, suspicious, anomaly, uncommon, unusual, rare, atypical and outliers. The definition of salient motion behavior has been causing much debate and confusion in the literature due to the subjective nature and complexity of human behaviors. In particular, they can be categorized into 2 broad understanding, where an event is considered salient if: i) There is deviation from the ordinary observed or learned events (i.e. the event having low occurrence or statistical representation in the learned model) ii) The event is not known or it is outstanding. 

\begin{figure}[!h]
\centering
\subfloat[Input sequence: Crowd in the marathon scene. The abnormal region (enclosed in the red bounding box) is simulated by inserting sythetic instabilities into the original videos.]{\includegraphics[height=0.25\linewidth, width=0.45\linewidth]{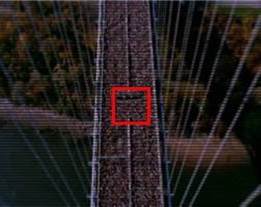}}
\qquad
\subfloat[Example abnormal region detected using the global motion saliencey detection method based on spectral analysis as proposed in \protect \cite{Loy12}. ]{\includegraphics[height=0.25\linewidth, width=0.45\linewidth]{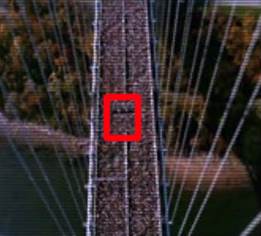}}
\qquad
\subfloat[Example abnormal region detected by exploiting the instability flow information as proposed in \protect \cite{Ali2007}.]{\includegraphics[height=0.25\linewidth, width=0.45\linewidth]{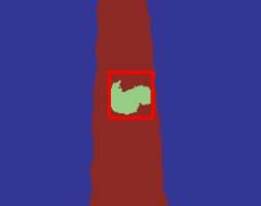}}
\qquad
\subfloat[Example output using the global similarity structure as proposed in \protect \cite{LimICPR2014} with the addition of discovering intrinsic structure of the motion dynamics, as illustrated in the coloured regions.]{\includegraphics[height=0.25\linewidth, width=0.45\linewidth]{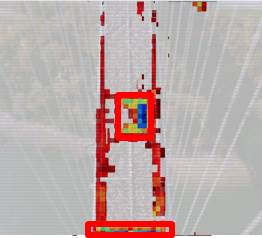}}
\caption{Comparisons between state-of-the-art anomaly detectors on the corrupted marathon sequence.}
\label{fig:Anomaly}
\end{figure}

Researchers have found that saliency can be identified and localized by exploiting the motion dynamics in crowded scene \cite{Ali2007,ChenNeuro2011,Loy12,Solmaz2012,LimICPR2014,zhu2014sparse} (examples as shown in Figure \ref{fig:Anomaly}). In particular, it has been observed that high motion dynamics and irregularities in the crowd motion are indeed good indicators of anomaly. Here, the high motion dynamics and irregularities constitutes to the emergent behavior. Lim \etal \cite{LimICPR2014}, proposed using the global similarity structure, which is a projection of the low-level representation of crowd motion to identify anomaly. Their experiments demonstrated that in dense crowd scenes such as the pilgrimage and marathon scenarios, the motions of individuals tend to follow the regular or dominant flow, resulting in stable motion dynamics. Thus, the possibility of anomalies taking place can be considered when there is high motion dynamics (unstable) and irregularities in a particular region. Some example scenarios of high motion dynamics and irregularities include stop-and-go waves, bottleneck and sources and sinks. This hypothesis again alludes to the aforementioned concept of collective motion, where individuals has high tendency to move with the crowd. In another variation in \cite{WangBayesian2007,WangBayesian2009}, the low-level representation are used to describe the `atomic' activities while the interactions between individuals are modeled for higher level understanding of crowd activity. Under the Bayesian model, their method detects saliency assigning marginal likelihood to the quantized motion. These methods are closer to the mesoscopic understanding of crowd, where the motion of individuals is combined with their interactions within the crowd to infer anomaly. Tackling salient motion behavior of crowd scene from a different aspect, Shao \etal \cite{shao2014scene} focuses on group profiles in crowd scene based on the notion that groups are the primary entities that constitute a crowd. The proposed framework aims at understanding crowd dynamics on a group level by using a fundamental set of intra and inter-group properties comprising collectiveness, stability, uniformity and conflict. There are also other approaches that adopt learning methods to interpret crowd dynamics for saliency detection.  Generally, large amount of data is required to enable good supervised/unsupervised learning for discriminative or generative crowd models. However, a major challenge in the context of crowd analysis in surveillance applications is the lack of abnormal or ground truth events for training. Andrade \etal \cite{Andrade2006} proposed an unsupervised method to model the degree of similarity between the trained model and new unseen video data. Their method applied spectral clustering on the flow field information to find the optimal number of models to represent normal motion dynamics, followed by a variation of Hidden Markov Model (HMM) for learning. Another variation in \cite{Kratz2009} proposed a 3D Gaussian distributions representation of spatio-temporal motion patterns which is then fed into a variant of HMM to discover the relationships between these patterns. Saliency is defined as statistical deviations within the video sequences of the same scene. In the more recent works in \cite{Mahadevan2010,LiPAMI2014},  a joint models of appearance and dynamics is proposed, known as the dynamic textures (DT).  Hierarchical mixtures of DT models are then performed, where the spatial and temporal saliency scores are integrated across time, space and scale with a conditional random field (CRF). Here, saliency is defined as events of low probability with respect to a model or normal crowd behavior. In \cite{Ihaddadene2008}, a non-learning method for crowd dynamic analysis is proposed, to mitigate the need of requiring a huge amount of data for accurate learning. Instead, their proposed method detects saliency by observing the deviations of features between a set of points-of-interest (POI) over a time series. In particular, the feature measurements include density, velocity and motion direction. Although the non-learning method provides convenient solution, it is restricted to a particular behavior or event such as detecting collapse flow near escalator exits and may not be ideal in dealing with the complexity of real-world scenarios. Similarly, a non-learning method based on threshold on kinetic energy of the crowd model is proposed to detect specific event such as running \cite{XiongNeuro2012}.  In addition, various unsupervised learning methods which suggests multi-level analysis (i.e. coarse-to-fine, global and local feature extractions) are proposed to deal with the complexity of crowd behavior \cite{XuNeuro2014,LiNeuro2015}.

\subsection{Crowd density estimation}


Not all events with large gathering of people are conducted in an enclosed venue with turnstiles where crowd density estimation can be administered seamlessly. And for some events such as parades or political protest, employing professionals to conduct human counting is infeasible. Nevertheless, estimating density of crowd is of utmost importance to better administer the well-being of crowd as a whole, development of public space design and accurate documentation of historical events. The Hillsborough disaster~\cite{LordTaylorHillsborough1989} is an example of the consequences of overcrowding.

\begin{figure}[!h]
\centering
\subfloat[Sample results of detecting individuals in crowd for density estimation on the USC dataset as presented in \protect \cite{rabaud2006counting}.]{\includegraphics[height=0.3\linewidth, width=1\linewidth]{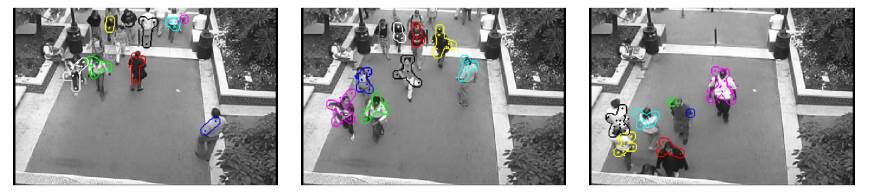}}
\qquad
\subfloat[Sample results on the PETS dataset, overlaid on the original images and foreground masks. The proposed method as in \protect \cite{Ge2010}, demonstrates promising estimation accuracy.]{\includegraphics[height=0.3\linewidth, width=0.98\linewidth]{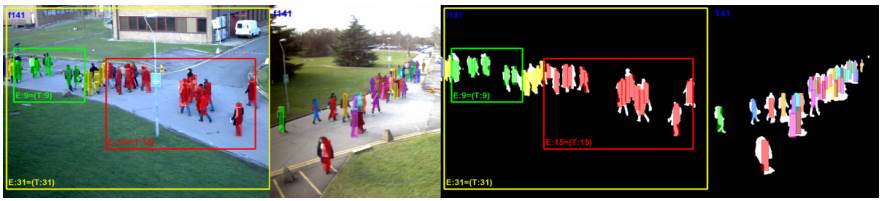}}
\caption{Sample outputs  of state-of-the-art methods for crowd density estimation.}
\label{fig:Density}
\end{figure}

On the contrary to the former two aspects of crowd behavior analysis, crowd density estimation is independent of the `thinking' component of each entity in crowd. Existing work on crowd density estimation depends mainly on collective motion and appearance cues, with respect to the type of inputs (i.e. crowd video sequences or single crowd image). Different techniques are adopted to cope with crowd scene of varying density. The greater density of crowd in a scene, the more complicated the task to estimate crowd density where dynamic occlusions come into picture. It is infeasible to discerned different person and ones' body parts when a person may only be occupying few pixels \cite{Haroon2013} and further rendered by background clutter. For instance, framework that performs clustering of coherent trajectories to represent a moving entity, and inferring number of individual in the scene by Rabaud and Belongie \cite{rabaud2006counting}, is limited to crowd scenes of sparse crowd where continuous sets of image frames are accessible.  The results presented in their work illustrated that for some crowd scene where individuals are closely positioned with each other, trajectories are incorrectly merged. This is due to the phenomenon of collective motion occurring between moving interacting entities. Using an analogous perception, Li \etal \cite{li2008estimating} estimate the numbers of people in crowd by implementing foreground segmentation and head-shoulder detection approach. The proposed method was intended to address stationary crowd, where subtle motions of individual is crucial and deeply relied on in defining foreground segments. Nonetheless, the proposed framework is susceptible to inter-occlusion between individuals, particularly prominent in a dense crowd scene. Ge and Collins in \cite{ge2009marked}, proposed a Bayesian marked point process to detect individuals in crowd where clear silhouette of individuals is required for accurate projection to a trained set for accurate detection and counting of individuals. In another study, Ge and Collins ~\cite{Ge2010} uses a generative sampling-based approach that leverage on multi-view geometry to achieve estimation of density of individuals in crowd. The work assumes that individuals in a crowd retain a certain space with each other (i.e. separation), which is one of the rules of interaction between entities in the crowd. Hence, individuals should not be occluded from all viewing angle. Example results of the aforementioned methods are shown in Figure \ref{fig:Density}.

Alleviating the need to detect each person in a crowd, some works \cite{marana1998automatic,davies1995crowd,Haroon2013,Chen2012,schofield1996system,tan2011semi,LiangNeuro2014} uses low level crowd features (appearance cue) formed based on the collectives of crowd to estimate crowd density. Marana et al. \cite{marana1998automatic} presented a method based on texture analysis to estimate crowd density, where the estimation is given in terms of discrete ranges (i.e. very low, low, moderate, high and very high). Their objective was to challenge scenes of dense crowd where each individual is greatly occluded.  They assumed that crowd scene of high density tend to illustrate fine textures, whereas crowd scene of low density are mostly made up of coarse patterns. Crowd density estimation by Davies et al. \cite{davies1995crowd} is one of the earliest works that uses regression approach to learn a linear relationship between low-level raw features (e.g. number of edge pixels) and crowd density. Similarly, works in \cite{chan2008privacy} and \cite{chan2012counting} propose to extract dynamic texture from homogeneous motion crowd segments and focus on learning mapping between large set of feature responses and density. A problem commonly encountered in regression based density estimation is perspective distortion, where individuals who are closer to the camera view appear larger than those who are positioned further away from the camera. The problem is exacerbated when single regression function is used for the whole image space. To address this problem, perspective normalization plays a key role by bringing the perceived size of individuals at different depths to the same scale. Another approach is to divide the image space into different cells and each cell is modeled by a regression function to mitigate the influence of perspective distortion. Idrees et al. \cite{Haroon2013} estimate the number of individuals given single dense crowd image by leveraging the harmonic textures elements of crowd from finer scales and appearances cues to approximate the density of crowd per image patch. The system uses regression approach to infer the count of individuals per patch and multi-scale random fields to refine the counts of individuals per image. Chen et al. \cite{Chen2012} proposed a multi-output regression approach to estimate crowd density in sparse crowd images. Low-level features extracted are shared among spatially localized regions to achieve more accurate counts prediction, indicating correlation between local regions of crowd scene is crucial. To compensate for insufficient and imbalanced training data inherent in regression approach, Zhang et al.~\cite{ZhangNeuro2015} proposed to utilize a label distribution learning method. Crowd images are annotated with label distributions, and thus can contribute to the learning of its real class and the neighboring classes. Consequently, training data for each class increases significantly. Chen et al. \cite{chen2013cumulative} introduced an attribute based crowd density estimation framework to address the data sparsity problem inherent in regression model. Low-level features extracted from image samples are mapped onto a cumulative attribute space using multi-output regression model to exploit the cumulative dependent nature between classes. Another regression model is learned to estimate crowd density using the attributes as input.

In another variations, Lempitsky and Zisserman \cite{lempitsky2010learning} model the density function over pixel grids, where integral over any region in the image would yield the density of object within. Kong et al. \cite{kong2006viewpoint} uses feed-forward neural network to map the correlation between feature histogram from low-level features and number of pedestrian. Line-of-interest (LOI) counting approach in \cite{cong2009flow} regards crowd motions as fluid flow to count number of individuals crossing the detection line within a time frame using flow velocity vector and dynamic mosaics. For a more comprehensive review of works in crowd density estimation, the reader is referred to \cite{loy2013crowd}. A complete taxonomy of the various approaches for crowd counting is provided and the key components of crowd density estimation framework are discussed in detail.

\section{The forthcoming crowd behavior analysis}
\label{Prospect}

There are several aspects of crowd behavior analysis which the authors believe are at their infancy and have the potential to develop further. \\

\noindent\textbf{Stationary crowd:} Crowds may essentially develop into two types, i.e., stationary or dynamic (moving) crowds. Stationary crowds are usually found as spectators or audiences at concerts, rallies, performances and speeches. Dynamic crowds are defined as crowd which is on the move, such as pilgrims that walk around the Kaaba during Hajj.

Most of the existing work on crowd focuses on moving patterns of individuals in the scene to infer their activities. Motion is often detected by using standard approaches such as frame-differencing to more complicated techniques such as dense optical flow. The estimated motion patterns are then analyzed to deduce various suggestions on the crowd activities. On the other hand, stationary crowd analysis has never been sufficiently investigated although the non-motion characteristics can provide rich information. This counter-intuitive approach of stationary crowd analysis is based on the notion that individuals or groups that remain in a particular area for a long time are worthy of attention. An earlier work by a well-known video analytics provider, the iOmniscient in \cite{iOmnscient2014}, provides a non-motion detection algorithm that has the ability to handle occlusion. The system is able to cope with hundreds of people moving around in a busy scene, to detect abandoned object as long as the object is visible for 50\% of the time. In a more advanced and recent work \cite{YiCVPR2014} and \cite{yi2014profiling}, a stationary crowd analysis method is proposed to detect four major activities; group gathering, stopping by, relocating and deforming. This work alludes to the findings of \cite{MoussaidPlos2010}, where their simulation on groups in crowd shows that stationary groups have greater impact on the dynamics of the scene than moving groups in some cases. This is justified further by simulating individuals forming stationary groups. The formation of stationary groups acts as an obstruction that changes the motion directions and dynamics of other individuals in the scene. Stationary crowd analysis is still at its early stage of research and is definitely worthy of upcoming investigations for a broader degree of scene understanding and traffic pattern analysis, in particular.\\

\noindent\textbf{Still images:} The study of literature in crowd behavior analysis found that most works are focused on video. Only limited areas of crowd behavior analysis are focused on using single image. Generally, the video-based crowd analysis captures motions of crowd over a duration or throughout a sequence of images. Meanwhile, image-based crowd analysis are based on a single image only, and the crowd can be stationary or dynamic. Although using video-based techniques have the advantage of utilizing temporal information, image-based methods allow separation between the appearance cues of crowd and background clutter. The appearance cues from image-based methods therefore, can be utilized to complement the temporal information from videos for an enhanced crowd behavior understanding. \\


\noindent\textbf{Large-scale:} 
Most recently, the computer vision field has witnessed a great leap forward through the adoption of deep learning neural networks to solve vision problems, including face recognition, image classification and pedestrian detection \cite{ChenBigData2014,ZengDeep2014,SunDeep2014}. One of the interesting features of deep learning is its capability to learn and train models based on large quantity of data, also known as the `Big Data'.  'Big Data' has earned various definitions across the different domains, and in this context it refers to the exponential growth and wide availability of digital data caused by the proliferation of CCTVs in public spaces nowadays. The explosion of videos on real-time public monitoring creates opportunities to utilize the large learning capacity of deep learning and models in the domain of video analysis. However, it is important to note that the conventional perception of using deep learning models as a black-box that is able to miraculously solve vision problems is reaching its plateau. Instead, there is a paradigm shift in the vision research community where vision problems are solved from the machine learning perspectives by casting them as high-dimensional data transform problems \cite{DongSuperResolution2014,LuoDeep2014}. We anticipate to witness the potentials of applying deep learning mechanisms into the context of crowd behavior analysis. 

Despite the robustness and flexibility of the deep learning architecture in learning an optimized feature representation automatically from the input data, it is highly dependent on the domain knowledge. This will lead to data-driven or application-driven discovery of more sophisticated deep learning models in the future. The application of deep learning to the field of video-based analysis is fairly new and amongst the earliest attempt of utilizing deep models for human crowds is presented in \cite{KangSegmentation2014}. In \cite{KangSegmentation2014}, the proposed framework has not fully utilized the motion information in video, as the motion filters are pre-trained before jointly optimized with the appearance filters. It would be interesting to see the evolution of the current deep learning framework to deal with the temporal or motion information in video sequences, in conjunction with the context of this review. Several other attempts at using deep learning architecture for crowd behavior analysis focuses on crowd scenes understanding \cite{shao2015deeply}  and crowd density estimation \cite{zhang2015cross}. The potentials of deep learning in transfer learning have yet to be fully realized in this field. The technical challenges imposed with this new way of looking at vision problems is relatively new and much more needed to be investigated in order to realized its potentials.

\section{Conclusion}
\label{Conclusion}

At this end, it is acknowledged that the precise resemblance or distinction between the physics and biological inspired approaches for crowd behavior analysis is rather vague. This is emphasized further by the complex nature of human behavior, especially in a crowd, and the various perceptions of it from the context of computer vision. Nevertheless, this survey aims to provide a platform within which, to address both the well explored and the neglected corners of these two sciences in the aspect of crowd behavior analysis, in particular. It is not the intention of this survey to take a stand on the implicit hierarchy of sciences which has long plagued the research scientists; with physics as the most respectable and biology as the conceptually poor cousin \cite{WinkelmanEnvy2008}, or the recent opinion that applauds biology~\cite{PennyReport2005,BribiescasBookReview2005}. Instead, this survey observes the paradigm shift that integrates the two sciences, to some degree, and the potentials of exploiting the two opinions to advance further the field of computer vision. This survey believes that most biological processes are governed by the laws of physics, yet it does not deny the progressive use of biological metaphors to understand problems in the various domains, from computer to physical systems. 

We suggest the need to bridge the different disciplines in coping with the exceedingly complex nature of human behavior. Understanding the interface between physics and biologically inspired algorithms is crucial towards developing `living' computer vision systems.  Recently, a new study of animal swarms has uncovered a new characteristic of their collective behavior when overcrowding sets in \cite{Romenskyy2013}.  This study was inspired by the condensed matter models, used for example in the study of magnetism in physics. In the reverse scenario that depicts how physics methods are inspired by biology includes introducing the concept of `thinking particles' for a higher level of fluid simulation \cite{sime1995crowd}. In this study, we merely described the direct and indirect connection between two main disciplines; the biology and physics domain. We found that they are indeed complimentary, although the fusion of both has yet to be fully utilized in the area of understanding and analyzing crowd behaviors. We reckon that the understanding of human behavior is far more complex and limitless to be modeled by these two approaches alone, and that there is a considerable potential for multidisciplinary areas involving statistical physics, sociology, philosophy and other branches of life sciences. In this survey, we raised questions for future research, to bridge the gap between current solutions and the real world requirements. From the continuously growing number of exciting new publications on physics and biologically inspired algorithms to solve computer vision, we conclude that this is indeed an emerging field that is worthy of future investigations.


\bibliographystyle{elsarticle-num}
\bibliography{myref}

\end{document}